\documentclass[fleqn,3p]{elsarticle}
\usepackage{graphicx}
\usepackage{amsmath,url}
\usepackage{multirow}
\usepackage{hhline}
\usepackage{rotating}
\usepackage{color, colortbl}
\usepackage{tabu}
\usepackage{xcolor}
\definecolor{Gray}{gray}{0.9}
\usepackage{amssymb}
\usepackage{array}
\newcolumntype{L}[1]{>{\raggedright\let\newline\\\arraybackslash\hspace{0pt}}m{#1}}
\newcolumntype{C}[1]{>{\centering\let\newline\\\arraybackslash\hspace{0pt}}m{#1}}
\newcolumntype{R}[1]{>{\raggedleft\let\newline\\\arraybackslash\hspace{0pt}}m{#1}}
\usepackage{color,soul}

\journal{Digital Signal Processing}

\begin{document}

\begin{frontmatter}

\title{Quality Measures for Speaker Verification with Short Utterances}


\author[mysecondaryaddress]{Arnab Poddar\corref{corr1}}
\ead{arnabpoddar@iitkgp.ac.in}
\address[mysecondaryaddress]{Department of Electronics \& Electrical Communication Engineering, \\ Indian Institute of Technology, India-721302, Kharagpur, India }
%

\author[mymainaddress]{Md Sahidullah}
\ead{md.sahidullah@inria.fr}

\author[mysecondaryaddress]{Goutam Saha}
\ead{gsaha@ece.iitkgp.ernet.in}

\address[mymainaddress]{MULTISPEECH Team, \\Universit\'{e} de Lorraine, CNRS, Inria, LORIA, F-54000, Nancy, France}
\cortext[corr1]{Corresponding author}

\begin{abstract}
The performances of the \emph{automatic speaker verification} (ASV) systems degrade due to the reduction in the amount of speech used for enrollment and verification. Combining multiple systems based on different features and classifiers considerably reduces speaker verification error rate with short utterances. This work attempts to incorporate supplementary information during the system combination process. We use quality of the estimated model parameters as supplementary information. We introduce a class of novel quality measures formulated using the zero-order sufficient statistics used during the i-vector extraction process. We have used the proposed quality measures as side information for combining ASV systems based on \emph{Gaussian mixture model-universal background model} (GMM-UBM) and i-vector. The proposed methods demonstrate considerable improvement in speaker recognition performance on NIST SRE corpora, especially in short duration conditions. We have also observed improvement over existing systems based on different duration-based quality measures.
\end{abstract}


\end{frontmatter}


\section{Introduction}

The \emph{automatic speaker verification} (ASV) technology uses the characteristics of human voice for the detection of individuals~\cite{kinnunen2010overview,campbell2009forensic}. The technology provides a low cost biometric solution suitable for real-world applications such as in banking~\cite{ICICI}, finance~\cite{Barclays}, and forensics~\cite{INTERPOL}. Similar to other traditional pattern recognition applications, an ASV system includes three fundamental modules~\cite{kinnunen2010overview,campbell1997speaker}: an \emph{acoustic feature extraction unit} that extracts relevant information from the speech signal in a compact manner, a \emph{modeling block} to represent those features and a \emph{scoring and decision} scheme to distinguish between genuine speakers and impostors. The state-of-the-art ASV system uses \emph{i-vector} technology that represents a speech utterance with a single vector of fixed length either using \emph{Gaussian mixture model-universal background model}~(GMM-UBM)~\cite{dehak2011front} or \emph{deep neural network}~(DNN)~  technology ~\cite{matvejka2016analysis}. More recently, deep neural network (DNN) based embeddings are used for speaker recognition~\cite{snyder2018xvector}. First, a DNN trained in a supervised manner to classify different speakers with known labels. Then, the trained DNN is employed to find a fixed-dimensional representation, known as \emph{x-vectors}~\cite{snyder2018xvector}, corresponding to a variable length speech utterance.

Despite of these recent technological advancements, the mismatch issues are still a major concern for its real-world applications \cite{poddar2017speaker}. The performance of ASV system considerably degrades in presence of mismatch due to \emph{intra-speaker variability} caused by the variations in speech duration~\cite{poddar2017speaker,solewicz2013estimated}, background noise~\cite{ming2007robust}, vocal effort~\cite{saeidi2016feature}, spoken languages~\cite{mclaren2016exploring}, emotion~\cite{parthasarathy2017study}, channels~\cite{wang2016robust}, room reverberation~\cite{vestman2017time}, etc.
In this paper, we focus on one of the most important mismatch factor, speech duration, the amount of speech data used in enrollment and verification.

\par


\subsection{Short utterance in speaker recognition}
State-of-the-art ASV systems exhibit satisfactory performance with adequately long (~2 minutes) speech data. However, reduction in amount of speech drastically degrades the ASV performance~\cite{poddar2017speaker,ming2007robust,kanagasundaram2011vector,mandasari2011evaluation,kanagasundaram2012plda}. The requirement of sufficiently long speech for training or testing, especially in presence of large intersession variability has limited the potential of widespread real-world implementations. An ASV system, in real world, is naturally constrained on the amount of speech data. Though this requirement can be fulfilled in training in some special cases, it is not always possible to maintain the same in verification for end-user convenience. In forensics applications, it is less likely to get sufficient data even for enrollment also~\cite{poddar2017speaker,mandasari2011evaluation}. Therefore, getting reliable performance for short duration speech is one of the most important requirement in ASV application.

\par
The performance of ASV systems are notably degraded with the reduction of amount of speech due to the lack of information provided in short utterance condition  \cite{mandasari2011evaluation,kanagasundaram2011vector,sarkar2012study,fauve2007influence}. In \cite{dehak2011front}, it is reported that the i-vector based ASV systems are less sensitive to limited duration utterances than \emph{support vector machine} (SVM) and JFA. The performance still deteriorates considerably with limited duration utterance as reported in \cite{kanagasundaram2012plda,kanagasundaram2011vector}. The duration variability problem is handled by extracting the duration pattern from the automatic speech recognition prior to modeling and scoring process in \cite{kajarekar2003modeling}. In \cite{fauve2008improving}, the short duration problem is approached, demonstrating the potential of fusion between GMM-UBM and SVM based systems using logistic regression. The work in \cite{hasan2013duration} attempted to model the duration variability as noise and also by a synthetic process. The work in \cite{kanagasundaram2014improving} has attempted to model variability caused by short duration segments in i-vector domain. In \cite{mandasari2015quality,mandasari2013quality}, i-vector based ASV system is calibrated for short duration using duration based quality measures.  The work in \cite{li2016improving} attempted to improve short utterance speaker recognition by modeling speech unit classes.

The latest DNN-based speaker embedding approaches have shown promising results for speaker recognition with short utterances~\cite{snyder2018xvector,zhang2018text}. Another recent work demonstrates that DNN-based i-vector mapping is useful for speaker recognition with short utterances~\cite{guo2018deep}. Even though the DNN-based methods give good recognition accuracy, they require massive amount of training data, careful selection of network architecture and related tuning parameters. In this current work, we aim at improving the speaker recognition performance by efficiently combining two popular ASV systems based on GMM-UBM and i-vector representation which require lesser number of tuning parameters and amount of training data compared to the DNN-based methods. Moreover, the GMM-UBM and i-vector method are suitable with limited computational resources.



\subsection{ Quality measure for duration-invariant speaker recognition}
  The research dealing with the effect of duration in speaker recognition have concentrated mostly on the consequences of classification performance, expressed in terms of \emph{equal error rate} (EER) and \emph{minimum detection cost function} (DCF) assuming the speaker model parameters are estimated satisfactorily.
However, the speaker models are affected due to duration variability in short duration. The idea of quality metric was successfully applied in biometric authentication systems \cite{poh2010quality,poh2005improving}. The quality metrics were employed to improve the efficiency of the multi-modal biometric systems~\cite{poh2010multimodal,poh2012unified,fierrez2005discriminative}. The work in \cite{grother2007performance} was motivated by a need to test claims that quality measures are predictive of matching performance. They also evaluated it by quantifying the association between estimated quality metric values and observed matching results.

The quality metrics are also successfully used in speech based
 bio-metric systems \cite{garcia2006using,garcia2004use}. The work in \cite{garcia2004use} studied a frame-level quality measure, obtaining encouraging results. However, the work in \cite{garcia2006using} showed a conventional user-independent multilevel SVM-based score fusion, adapted for the inclusion of quality information in the fusion process.
   The work in \cite{hasan2013crss} focused on quality measure based system fusion, giving the emphasis on noisy and short duration test conditions using NIST 2012 database.
 The commonly used ASV systems, such as i-vector and GMM-UBM, do not include the information about the quality of estimated speaker models and information of duration variability. The work documented in \cite{harriero2009analysis}, analyzed several quality measures for speaker verification from the point of  view of their utility in an authentication task by selecting several quality measures derived from classic indicator like ITU P.563 estimator of subjective quality, signal to noise ratio and kurtosis of linear predictive  coefficients. Moreover, the work \cite{harriero2009analysis} proposed a novel quality measure derived from what we have  called universal background model likelihood (UBML). The work in \cite{alonso2012quality}, analyzed the factors that negatively impact the biometric quality and also depict a review of overall framework for the challenges of biometric quality.

The work in \cite{mandasari2013quality} used duration of speech segments to formulate the quality metrics and subsequently utilized the same for the calibration of recognition scores. However, the duration based quality metrics may not improve performance where the duration is fixed for either enrollment or verification or both. These duration based quality measures ignored the information of quality of speaker-model estimation. The quality of speaker-model parameters are not only dependent on duration, noise but also on phonetic distribution, intelligibility of speech etc.  However, to develop a solution by targeting the basic building blocks of an ASV system, we attempted to incorporate the information of duration variability which degrades the quality of speaker-models. The concept of quality may be defined as degree of goodness of an element \cite{garcia2004use,garcia2006using}, which, in our case, is the speaker-models. We treat BW statistics not only as the source of speaker information but also as a source of quality of estimated speaker models.

The Baum-Welch (BW) statistics, which represent the speech features in the intermediate step of i-vector extraction process, is affected by the duration variability. Consequently, the variability gets propagated in the subsequent representation, i.e, i-vector. We hypothesize that BW statistics can help to extract the quality of speaker-model parameters. We demonstrate through graphical analysis that the utterance duration is associated with the dissimilarity measures between intermediate statistics and background model parameters. We propose to use this measure as a quality measure. In this work, we propose to formulate this quality measure from the BW statistics and universal background model (UBM) parameters.

\begin{figure}[t]
\centering
\includegraphics[width=12cm]{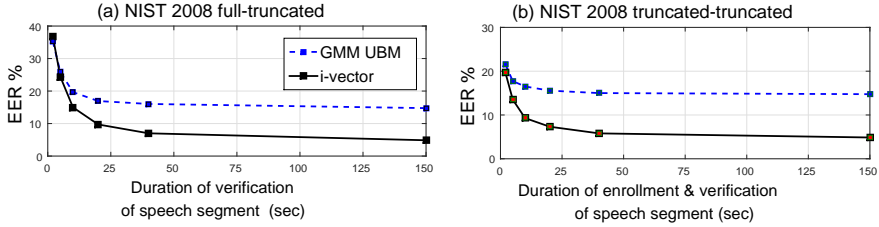}
\caption{Plot of speaker recognition performance on (a) NIST 2008 corpus (short2-short3) with truncated enrollment and truncated verification. (b) NIST 2008 short2-short3 corpus with full duration enrollment and truncated verification.}
\label{systemeer}

\end{figure}

The proposed quality measures can be infused as additional information in the ASV technique to improve the system performance. The quality measures can be incorporated in potentially four possible stages of ASV system: feature extraction, speaker-model training, score computation and fusion of scores \cite{garcia2006using}. The use of quality measures in score fusion stage is most straightforward and has been successfully applied in speech, finger-print, face based multimodal person authentication systems \cite{chibelushi2002review,kittler2007quality}. In this paper, we incorporate the proposed quality measures in score fusion stage to improve the performance of speaker recognition system in various duration conditions.
In short duration, the linear score fusion strategy showed efficient performance with GMM and SVM based classification framework~\cite{fauve2008improving}. However, the i-vector based system (with GPLDA based channel compensation) was reported to perform more efficiently over JFA and GMM-SVM based framework in short utterance conditions~\cite{dehak2011front}. Here, we show a comparative performance study of i-vector and GMM-UBM on NIST corpora (Fig. \ref{systemeer}). We observe that though i-vector system performs better than GMM-UBM for long duration speech, the GMM-UBM system still shows comparable or even better performance for short duration conditions~\cite{arnab2017icapr,poddar2018improved}. This observation inspire to fuse i-vector and GMM-UBM to develop a more accurate and reliable solution for practical application of ASV systems. We have incorporated  the estimated quality measures while blending the GMM-UBM and i-vector based ASV system. Incorporation of quality measures not only showed considerable improvement in performance but also consistency in various duration conditions. The proposed systems showed more relative improvement in short duration conditions which is more relevant for practical requirement. A preliminary version of this work was presented in~\cite{arnab2017icapr}. In this work, we conduct extensive analysis and experiments.


 \par
\par


    \par


\par
The rest of the paper is organized as follows. The theoretical aspects of classical GMM-UBM and i-vector GPLDA system are discussed in Section \ref{Theory}. Analysis on intermediate subsystems under different duration variability condition is presented in Section \ref{SUC}. Section \ref{ProposedQuality} describes the proposed quality measures and quality aided fusion based system.
Details of experimental setup are provided in Section \ref{Exp_set}. The comparison of performance metrics of baseline GMM-UBM and i-vector GPLDA based ASV system and results on proposed quality aided fusion system are reported in Section \ref{RnD}. Finally, conclusion is drawn in Section \ref{conclude}.


\section{Automatic Speaker Recognition System}
\label{Theory}
Speaker recognition system, based on Gaussian mixture model, has emerged as the most widely used fundamental approach with the introduction of universal background model \cite{reynolds2000speaker}. Subsequently, GMM supervector based SVM \cite{campbell2006svm}, and JFA \cite{kenny2007joint} were introduced in ASV technology. Recent state-of-the-art speaker recognition concentrates on compact representation of GMM supervectors, named as i-vectors \cite{dehak2011front}. This work considers ASV system based on subspace modeling of i-vectors using PLDA \cite{kenny2010bayesian}. This section presents a brief explanation of GMM-UBM, i-vector and PLDA.
\subsection{GMM-UBM based ASV system}\label{section1}

In GMM-UBM, prior to enrollment phase, a single speaker independent universal background model is created by using  a large development data \cite{reynolds2000speaker}. The UBM is represented as $\boldsymbol{\lambda}_{\mathrm{UBM}}=\{ w_{i},\boldsymbol{\mu}_{i},\mathbf{\Sigma}_{i} \}_{i=1}^{C}$ where $C$ is the total number of Gaussian mixture components, $w_{i}$ is the weight or prior of $i-th$ mixture component, $ \boldsymbol{\mu}_{i}$ is the mean and $\mathbf{\Sigma}_{i}$ represents the co-variance matrix. Parameter $w_{i}$ satisfies the constrain $\displaystyle \sum_{i=1}^{C}w_{i}=1$.
\par
A group of $S$ speakers is represented by their corresponding model as $ \{ \boldsymbol{\lambda}_{1},\boldsymbol{\lambda}_{2},\ldots, \boldsymbol{\lambda}_{S} \} $. In the GMM-UBM system, we derive the target speaker model by adapting the GMM-UBM parameters. The model parameters are adapted by \emph{maximum-a-posteriori} (MAP) method. First, sufficient statistics $ N_{i} $ and  $\mathbf{E}_{i}$ from a hypothesised speaker's utterance with $T$ speech frames $\mathbf{X} =  \{ \mathbf{x}_{1}, \mathbf{x}_{2},\ldots, \mathbf{x}_{T} \}$,  are calculated as,
\begin{equation}
\label{eq:ni}
  N_{i}= \displaystyle \sum_{t=1}^{T}Pr(i|\mathbf{x}_{t})
   \quad \mathrm{and} \quad \mathbf{E}_{i}(\mathbf{X})= \frac{1}{N_{i}} \sum_{t=1}^{T}Pr(i|\mathbf{x}_{t})\mathbf{x}_{t}
\end{equation}
where posterior probability of i-th component  $Pr(i|\mathbf{x}_{t})$, is conditioned on speech data $\mathbf{X} =  \{ \mathbf{x}_{1}, \mathbf{x}_{2},\ldots, \mathbf{x}_{T} \}$.

\par
In the testing phase, average log-likelihood ratio $\Lambda(\mathbf{X}^{\mathrm{test}})$ is determined using test feature vector $ \mathbf{X}^{\mathrm{test}}= \{\mathbf{x}_{1}, \mathbf{x}_{2}, \ldots, \mathbf{x}_{T,\mathrm{test}} \} $ against both target model and the background model.
\begin{equation}
\Lambda_{\mathrm{UBM}}(\mathbf{X}^{\mathrm{test}})= \log\ p(\mathbf{X}^{\mathrm{test}}|\boldsymbol{\lambda}_{\mathrm{target}})- \log\ p(\mathbf{X}^{\mathrm{test}}|\boldsymbol{\lambda}_{\mathrm{UBM}})
\end{equation}
Finally, a decision logic is applied to decide whether the claimant speaker will be accepted or rejected. A decision threshold is used for decision, like if
$  \Lambda_{\mathrm{UBM}}(\mathbf{X}^{\mathrm{test}})$ exceeds a predefined threshold
 then the claim will be accepted, else rejected.


\subsection{i-vector based ASV system}\label{TVs}
The i-vector represents the GMM supervector in a \emph{total variability}  space which reduces dimension of GMM supervector  \cite{dehak2011front}. In i-vector space, the GMM supervector, i.e, the concatenated means of GMM mixture components, is represented as
$  \mathbf{M}=\mathbf{m}+\mathbf{\Phi y}
$,
where $\mathbf{\Phi}$ is a low-rank total variability matrix and $\mathbf{y}$ represents i-vector, $\mathbf{m}$ is the speaker and channel independent supervector (taken to be UBM supervector) and $\mathbf{M}$ is the speaker and channel dependent GMM supervector.

Zeroth and first order BW statistics
$N_i$ and $\mathbf{E}_i$ respectively are used to obtain the i-vector $\mathbf{y}$. The prior distribution of i-vectors $p(\mathbf{y})$ is assumed to be $ \mathcal{N}(0,I)$ and posterior distribution of $\mathbf{E}$, conditioned on the i-vector $\mathbf{y}$ is hypothesized to be $p(\mathbf{F|y})=\mathcal{N}(\mathbf{\Phi} \mathbf{y}, \mathbf{N}^{-1}\mathbf{\Sigma})$. The MAP estimate of $\mathbf{y}$ conditioned on $\mathbf{E}$ is given by
\begin{equation}
   \mathbb{E}(\mathbf{y|E})=(\mathbf{I}+ \mathbf{\Phi}^{\top} \mathbf{\Sigma}^{-1}\mathbf{N\Phi)^{-1}\Phi}^{\top}\mathbf{\Sigma}^{-1}\mathbf{N}(\mathbf{E}-\bar{\mathbf{m}}),
\end{equation}

where $\mathbb{E}(\mathbf{y|E})$, the expected value of the posterior distribution of $\mathbf{y}$ conditioned on $\mathbf{E}$ is considered as the \emph{i-vector} representation of a speech utterance. Here $\mathbf{I}$ refers to Identity matrix, the term $\mathbf{\Phi}$
 refers to total variability matrix, estimated from the development data. The symbol $\mathbf{\Sigma}$ refers the co-variance matrix adopted from UBM. The term $\bar{\mathbf{m}}$ represents the concatenated mean of the UBM components.

\subsection{Gaussian Probabilistic Linear Discriminate Analysis (GPLDA)}
A recent attempt to model speaker and channel variability in i-vector space is accomplished through probabilistic LDA (PLDA) modeling approach. In this paper, we have used a simplified variant of PLDA, named as Gaussian PLDA \cite{kenny2010bayesian}. The inter-speaker variability is modeled by a full co-variance residual term. The generative model for $s$-th speaker and $j$-th recording of new i-vector variability projected space is given by
\begin{equation}
\mathbf{y}_{s,j}=\boldsymbol{\eta}+\mathbf{\Psi z}_s+\boldsymbol{\epsilon}_{s,j},
\end{equation}
where $\boldsymbol{\eta}$ is the mean of the development i-vectors, $\mathbf{\Psi}$ is eigen-voice subspace and $\mathbf{z}$ is a vector of latent factors. The residual term $\boldsymbol{\epsilon}$ represents the variability not captured by the latent variables. This generative model approach of i-vector space representation has been applied successfully with considerable improvement in recognition accuracy \cite{kenny2010bayesian}.
\subsection{Likelihood Computation}
score calculation of GPLDA based i-vector system  uses likelihood ratio \cite{kenny2010bayesian}. For a projected enrollment and verification i-vector $\mathbf{z}_{\mathrm{target}}$ and $\mathbf{z}_{\mathrm{test}}$ respectively, the likelihood ratio $\Lambda_{\mathrm{GPLDA}}( \mathbf{z}_{\mathrm{target}}, \mathbf{z}_{\mathrm{test}} )$ can be calculated as follows:

\begin{equation}
  \mathbf{\Lambda}_{\mathrm{GPLDA}}(\mathbf{z}_{\mathrm{target}}, \mathbf{z}_{\mathrm{test}})=\log\ \frac{ \space p(\mathbf{z}_{\mathrm{target}}, \mathbf{z}_{\mathrm{test}}|H_1)} { p(\mathbf{z}_{\mathrm{target}}|H_0)\ p(\mathbf{z}_{\mathrm{test}}|H_0)}
\end{equation}
where $H_0$:  The i-vectors belong to different speaker.\\ $H_1$:  The i-vectors belong to the same speaker.



\begin{figure*}[t]
\centering
 \includegraphics[width=10cm]{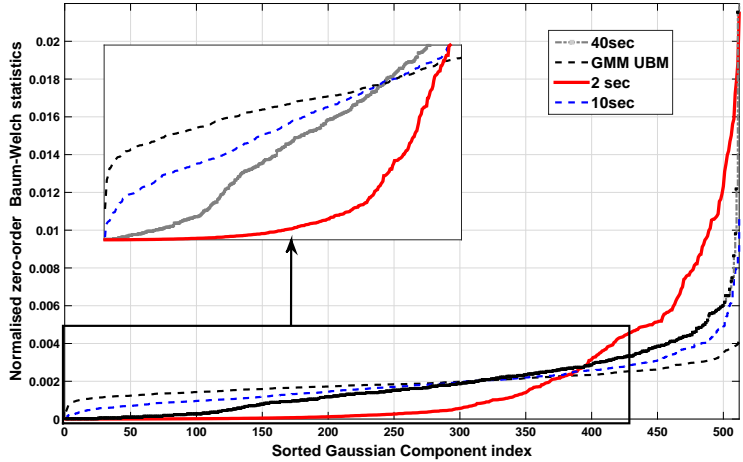}
\caption{ Sorted distribution of normalized zeroth-order Baum–Welch statistics $(\tilde{N}_i)$ corresponding to various duration conditions (2~sec, 10~sec, 40~sec). Prior weights of GMM-UBM components $(w_i)$ are also shown in sorted ascending order.}

\label{NscatterFull}
\end{figure*}

\section{Analysis on BW statistics extraction procedure}
\label{SUC}

Previous studies dealing with duration variability concentrated on the final performance metrics measured in terms of EER, DCF, etc \cite{kanagasundaram2011vector}. Some studies focused the variability in i-vector space  \cite{kanagasundaram2014improving}.
However in this work, we present a study on how duration variability affects the intermediate steps of ASV system.
BW statistics represent the total information from the speech and are transformed into i-vectors for decision making. Since, in most of modern ASV systems, BW statistics is an indispensable and important step, we initially conduct analysis on the BW sufficient statistics to investigate the characteristics and effect of the short duration.

We have the relationship for zeroth order BW statistics as,
$N_i= \displaystyle \sum_{t=1}^{T}Pr(i|\mathbf{x}_{t})$, summing over all Gaussian mixture components $C$ we obtain,
\begin{equation}
\sum_{i=1}^{C}N_{i}= \displaystyle \sum_{i=1}^{C}\sum_{t=1}^{T}Pr(i|\mathbf{x}_{t}) = T
\end{equation}
Normalizing zeroth order statistics for a single Gaussian mixture component $i$ we get
\vspace{-2mm}
\begin{equation}
\label{ni}
\tilde{N}_i= \displaystyle \frac{1}{T}\sum_{t=1}^{T}Pr(i|\mathbf{x}_{t}) \quad
\mathrm{and} \quad \sum_{i=1}^{C}\tilde{N}_i=1
\end{equation}
Hence normalized zeroth order BW statistics (NBS) has the same property as weights of the GMM-UBM, i.e., $\sum_{i=1}^{C}w_i=1 $ . Moreover, $\tilde{N}_i$ can be regarded as the mixture weight indicator of the Gaussian component $i$ for a particular speech segment.
  It is a standard statistical hypothesis that BW statistics are better estimated with sufficiently large speech data which capture all kinds of variability with meaningful proportion. Hence, it is expected that the higher value of $T$ with phonetically rich speech segment would lead to better quality of estimation of $\tilde{N}_i$. On the other hand, the intermediate statistics may be expected to be updated more sparsely for reduced speech data or degraded quality of speech.
  On this core note, the characteristics of $\tilde{N}_i$ are investigated. Systematic studies are presented separately on single utterance and multiple utterances from multiple speakers.

\subsection{Baum-Welch statistics for short-utterance}
Initially, a telephone utterance from NIST SRE 2008 short-2 enrollment corpus of male speakers is taken for analysis on NBS of different duration conditions, e.g., 2 sec, 10 sec and full length (1.5 - 2.5 mins).
In Fig. \ref{NscatterFull}, $\tilde{N}$ is plotted in ascending order with respect to Gaussian mixture components for different duration conditions.
We observe from Fig. \ref{NscatterFull}  that the gradient is steeper for short duration compared to the other longer duration conditions, whereas that of the UBM showed more flat nature. This observation refers to the fact that only a few number of Gaussian components are associated with most of the speech frames in limited duration conditions. A large number of Gaussian components do not associates adequate speaker-specific frames which finally affects the quality of model estimation. This effect reduces as the duration of the utterance increases. As GMM-UBM is estimated from sufficiently large pool of speech data, most of the Gaussian components are occupied by adequate speech frames. Thus, it shows more uniform nature in Fig. \ref{NscatterFull}. We hypothesize that the
introduction of greater variability in zeroth order statistics  especially for short duration condition indicates the lower quality of estimation of model parameters (NBS). This indicates the quality of estimation of NBS is degraded in short duration condition. We treat NBS not only as a source of speaker information, it can help measure the quality of estimated model of speakers.


 \par



\begin{figure}[t]
\centerline{
\includegraphics[width=17cm]{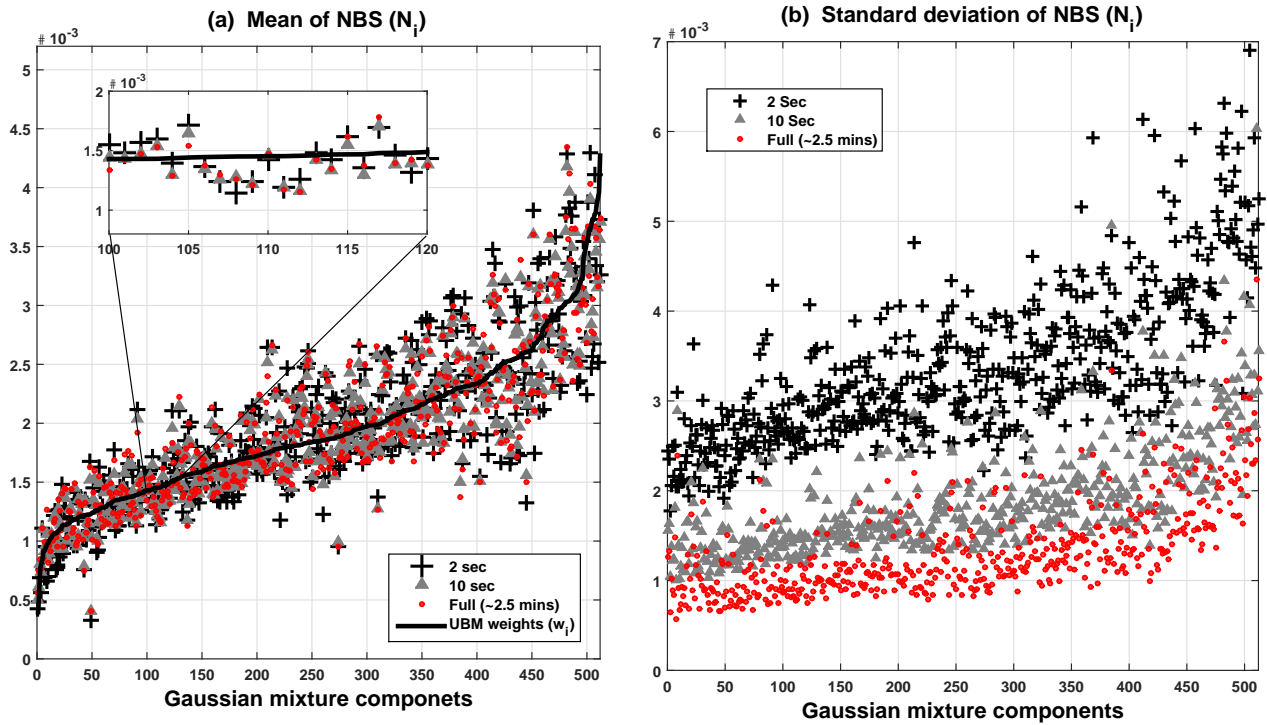}}

 \caption{(a)Mean and (b)standard deviation of normalized Baum-Welch statistics (NBS) for each of the 512 Gaussian components computed on three duration conditions (2 Sec, 10 Sec and full). Figure(a) includes the weights of GMM-UBM for corresponding Gaussian. We have used all the male speakers of NIST 2008 (short2) corpus for this analysis.}
\label{NdistUBM}

\end{figure}

\subsection{Analysis of NBS in multiple speech utterances}

In order to interpret the NBS, we have done an analysis on entire male part of NIST SRE 2008 consisting. The effect of speech duration on the NBS is presented in Fig.~\ref{NdistUBM} by showing mean (Fig.~\ref{NdistUBM}(a)) and standard deviation(Fig.~\ref{NdistUBM}(b)) of NBS per Gaussian components for three duration conditions (2 sec, 10 sec and full length). We have also shown the weights of GMM-UBM of corresponding mixture components $(w_i)$ are presented in Fig. \ref{NdistUBM}(a).

We observe in  Fig. \ref{NdistUBM}(a) that the means of $\tilde{N}$ for different duration condition follows the value of UBM weight of corresponding Gaussian mixture component $(w_i)$. We also notice that for different duration conditions, the means of different duration condition for a particular Gaussian component remains nearly similar. This is observed in almost all Gaussian components shown in Fig. \ref{NdistUBM}(a). These observations on means of $\tilde{N}$ distribution and weights of corresponding Gaussian mixture component inspired us to use GMM-UBM weights $(w_i)$ as reference to measure the variability in $\tilde{N}$.

The short segments in Fig. \ref{NdistUBM}(b) also show greater standard deviation referring greater variability introduced in NBS. We observe gradual increment in the variability with a reduction in speech segment length. As the variability in NBS is due to the duration, we hypothesize that the duration-related quality can be reflected in the NBS.

\section{Quality measures for speech segments}
\label{ProposedQuality}
The observations in previous section illustrate that the variability in BW statistics is in someway associated with the duration of speech. The work in~\cite{li2016feature} also attempted to model the sparsity and variability to compensate the performance of ASV in short duration. The work in~\cite{poorjam2016incorporating} introduced an uncertainty measure computed from the the i-vector posterior parameter to compensate the duration variability effect.
In our current work, we propose to apply dissimilarity metrics between NBS ($\tilde{N}_i$) and prior of corresponding Gaussian component of UBM model ($w_i$) to measure the impact of duration on speaker representation. Subsequently, this is incorporated as supporting information in fusion of ASV system.

\begin{table}[t]
 \caption{Mathematical expression for quality measures formulated using normalized Baum-Welch statistics and UBM weights.}
 \centerline{
 \setlength{\tabcolsep}{5pt}
\label{qmf}       
  \begin{tabular}{lc}
\hline\noalign{\smallskip}\noalign{\smallskip}
 Quality Measure & Short form\\
\noalign{\smallskip}\hline\noalign{\smallskip}\noalign{\smallskip}
 $Q_1(\tilde{N}_s)=\sum^{C}_{i=1}(\tilde{N}_{i,s}log\frac{\tilde{N}_{i,s}}{w_{i,\mathrm{ubm}}})$ & kl-1\\
\noalign{\smallskip}\noalign{\smallskip}
 $Q_2(\tilde{N}_s)=\sum^{C}_{i=1}(w_{i,\mathrm{ubm}}log\frac{w_{i,\mathrm{ubm}}}{\tilde{N}_{i,s}})$ & kl-2\\
\noalign{\smallskip}\noalign{\smallskip}
 $Q_3(\tilde{N}_s)=\frac{1}{2}\sum^{C}_{i=1}(\tilde{N}_{i,s}log\frac{\tilde{N}_{i,s}}{w_{i,\mathrm{ubm}}}+w_{i,\mathrm{ubm}}log\frac{w_{i,\mathrm{ubm}}}{\tilde{N}_{i,s}})$ & kl-avg\\
\noalign{\smallskip}\noalign{\smallskip}
 $Q_4(\tilde{N}_s)=\sum^{C}_{i=1}|\tilde{N}_{i,s}-w_{i,\mathrm{ubm}}|$ & $\ell_1$ norm\\
\noalign{\smallskip}\noalign{\smallskip}
$Q_5(\tilde{N}_s)=\sqrt{(\sum^{C}_{i=1}|\tilde{N}_{i,s}-w_{i,\mathrm{ubm}}}|^{2}$ & $\ell_2$ norm\\
\noalign{\smallskip}\noalign{\smallskip}
$Q_6(\tilde{N}_s)=\sqrt{(\sum^{C}_{i=1}\tilde{N}_{i,s}w_{i,\mathrm{ubm}})}$ & bh\\
\noalign{\smallskip}\noalign{\smallskip}
\noalign{\smallskip}\hline
\end{tabular}
}

\end{table}

\subsection{Quality Measure Modeling}
The trends observed in Section \ref{SUC}, are exploited to empirically model the quality of speech segments. In this paper, six types of dissimilarity measures are adopted to model the duration variability which degrades quality of speaker model estimation. The mathematical expressions to model the quality $Q$ of a segment $s$ are presented in Table \ref{qmf}. The adopted quality measures differ from the way of measuring dissimilarity of $\tilde{N}_i$  from the weights of UBM $(w_i)$, which is treated as reference. Quality measure operators $Q_1, Q_2$ and $ Q_3$ models the Kullback-Liebler divergence between  NBS $(\tilde{N}_{i,s})$ and weights of Gaussian mixture components of UBM $w_{\mathrm{i,\mathrm{ubm}}}$. These metrics attempts to capture the divergence of the distribution of NBS and UBM weights. Furthermore, we have also used other metrics to measure the dissimilarity. However, $\ell_1$-norm and $\ell_2$-norm are applied in quality measure operator $ Q_{4} and Q_{5}$ respectively. The Bhattacharyya distance is used in to measure the overlap of the population samples of NBS and weights of UBM.

\subsection{Statistical analysis of Quality Measures}
An analysis for statistical relevance of the modeling of quality measures is illustrated in this section. Segments from NIST 2008 short2 enrollment corpus with 1270 male speakers are used for assessment. The distribution mean of dissimilarity measures $\bar{Q}_j$ of all 1270 segments and its truncated versions (2~sec, 5~sec, 10~sec, 20~sec, full) are presented in Table \ref{qan}. The mathematical expression for the mean of quality measures of type $\bar{Q}_j$ estimated from NIST 2008 short-2 corpus consisting of $H$ utterances is given by
   \begin{equation}
   \bar{Q}_j=\frac{1}{H}\sum_{s=1}^{H}Q_{j}(\tilde{N}_s) \qquad j= 1, 2, \dots 6
   \vspace{2mm}
   \end {equation}

\begin{figure}[!t]
\centering
\includegraphics[width=12cm]{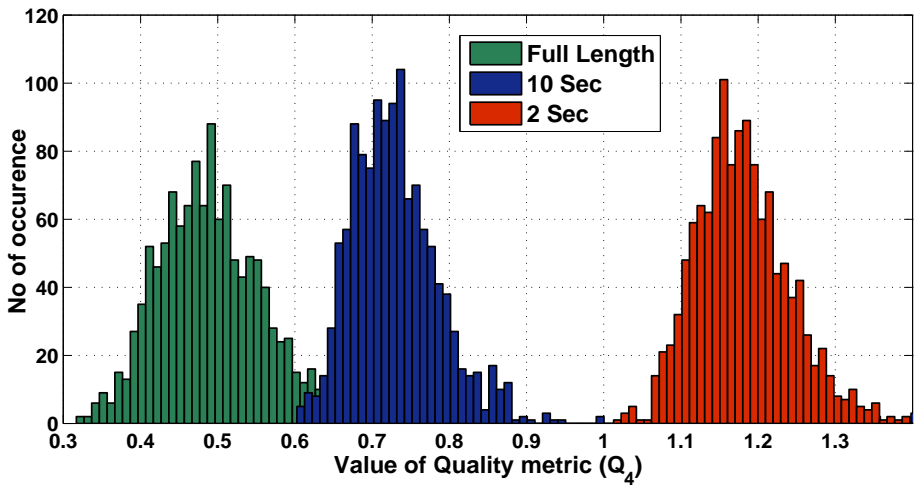}
\caption{Distribution of quality measure $Q_4$ for \emph{2~sec, 10~sec, and full} segment, estimated from \emph{NIST 2008 short2 enrollment} corpus.}
\label{QualDist}

\end{figure}

Separate analysis are presented in Table \ref{qan} for six types of dissimilarity measures $(Q_j, \quad j=1,2,\dots6)$ .
The distribution of quality measure of $Q_4$  for truncated version of 2 sec, 10 sec and full duration are plotted in Fig.\ref{QualDist}. Results indicate that $\bar{Q}_j$ is decreased as the duration of speech segment rises as shown in Table \ref{qan} and Fig. \ref{QualDist}. Hence, 2 sec segment shows highest value for $\bar{Q}_j$ and full duration segment shows the lowest. In Table \ref{qan}, we observe gradual decrements in distribution mean of dissimilarity measure $\bar{Q}_j$  with the increment of speech duration.  The decrements are consistent for almost all kind dissimilarity measures presented in Table \ref{qmf}. These observations leads to use the proposed dissimilarity measure as the overall quality of speaker model.

\begin{table}[t]
\caption{Mean of proposed quality measures of all segments in NIST 2008 short2 enrollment corpus and that of its truncated versions for all six types quality measures.}
 \setlength{\tabcolsep}{5pt}
\centering
\label{qan}       
  \begin{tabular}{lcccccc}
\hline\noalign{\smallskip}
duration & $\bar{Q_1}$ & $\bar{Q_2}$ &$\bar{Q_3}$&$\bar{Q_4}$&$\bar{Q_5}$&$\bar{Q_6}$\\
& kl-1 & kl-2 & kl-avg & $\ell_1$ norm & $\ell_2$ norm & bh\\
\noalign{\smallskip}\hline\noalign{\smallskip}
2 sec & 2.250 & 1.051 & 1.652 & 1.144 & 0.084 & 0.536\\
5 sec & 1.314 & 0.679 & 0.997 & 0.899 & 0.061  & 0.399\\
10 sec& 0.963 & 0.521 & 0.749 & 0.742 & 0.055  & 0.327\\
20 sec& 0.782 & 0.437 & 0.608 & 0.654 & 0.050  & 0.273\\
Full  & 0.309 & 0.226 & 0.268 & 0.499 & 0.031  & 0.178\\
\noalign{\smallskip}\hline
\end{tabular}

\end{table}

\subsection{Fusion of ASV systems with quality measures}

Quality measures can be incorporated in different stages of a speaker verification system like model training~\cite{ferrer2005class}, computation of scores \cite{mandasari2013quality,mandasari2015quality}, scores fusion \cite{doddington2000nist}, etc. The existing literature show that the quality information can be used in many applications, especially for score fusion in biometric person recognition system~\cite{chibelushi2002review,kittler2007quality,bigun2003multimodal,fierrez2005discriminative}. In our work also, we incorporate the quality measures in score fusion step. In ASV, fusion-based approaches have found very much suitable for improving recognition accuracy~\cite{hasan2013crss,hautamaki2013sparse}. Though i-vector and GMM-UBM based ASV systems uses different approaches, we show in Section~\ref{RnD} that GMM-UBM exhibit competitive or even better performance over i-vector in short utterance conditions.
We fuse i-vector and GMM-UBM to exploit the information captured simultaneously by two systems. The fusion parameters are trained using logistic regression objective using the BOSARIS toolkit~\cite{bosaris2011}. Separate NIST corpora, SRE 2008 \cite{NIST2008} and SRE 2010 \cite{NIST2010}, are used for training and evaluation of the fusion parameters respectively.

We confine our work to score level fusion with fusion function $f$ which combines two base classifier scores, $\Lambda_{UBM}$ and $\Lambda_{GPLDA}$, into a single score,
\begin{equation}
\mathbf{\Lambda}=\{\Lambda_{UBM},\Lambda_{GPLDA}\}^{\top}.
\end{equation}

 The decision is made by comparing the fused score to a threshold. An annotated development set $\mathcal{D}=\{(\mathbf{\Lambda}_{i},c_{i}), i= 1,2,\dots,N_{dev}\}$ with $c_{i}\in \{ +1,-1\}$ representing corresponding speech frame from target speaker $(c_{i}\in \{ +1\})$ or imposter $(c_{i} \in\ \{ -1\})$ is used to train the fusion parameters.

The general model of linear fusion of the two systems is represented by:
\begin{equation}
\label{qeqn_lin}
f_{\mathrm{lin}}(\mathbf{\Lambda})= \mathbf{\alpha}^{\top}\mathbf{\Lambda}+\bar{\theta} ,
\end{equation}

where $\mathbf{\alpha}$ is the fusion weight and $\theta$ is the bias. These fusion parameters are estimated by logistic regression on the development scores.

After incorporating the quality measures for enrollment ($Q(\tilde{N}_{\mathrm{enrol}})$) and test ($Q(\tilde{N}_{\mathrm{test}})$) utterances, the general model of fusion is given by:
\begin{equation}
\label{qeqn}
f_{Q}(\mathbf{\Lambda})= \mathbf{\alpha}_{Q}^{\top}\mathbf{\Lambda}+\theta_{Q} + \beta \times Q(\tilde{N}_{\mathrm{enrol}}) Q(\tilde{N}_{\mathrm{test}}),
\end{equation}

where $\mathbf{\alpha}_Q$, $\theta_{Q}$, and $\beta$ are the parameters for quality measure fusion. Note that quality-based fusion is also a type of linear fusion where the quality parameters are incorporated as additional similar scores. We have also conducted a separate experiment where we have incorporated the proposed quality metric in single i-vector system for duration-based score calibration~\cite{mandasari2013quality}. The general model for adding quality measure in single i-vector based system is given by,

\begin{equation}
\label{qeqn_ivec}
f_{C}(\mathbf{\Lambda})= \alpha_{C}\Lambda_{GPLDA}+\theta_{C} + \beta_C \times Q(\tilde{N}_{\mathrm{enrol})} Q(\tilde{N}_{\mathrm{test}})
\end{equation}

where the parameters $\alpha_{C}$, $\theta_{C}$ and $\beta_C$ are the parameters for calibration and they are estimated on the development set with \textit{known} labels and applied evaluation set with \textit{unknown} labels. In our experiments, we have observed that the i-vector based scores and GMM-UBM based scores can be fused linearly which yields encouraging performance improvement \cite{poddar2018improved,arnab2017icapr,arnab2015comparison}. Hence, we attempt to add quality information in linear fusion of i-vector and GMM-UBM system. However, while defining fusion model, we have incorporated the quality information in score fusion using trial-by-trial manner. We have attempted to model the quality of a trial by simply multiplying the quality measures, separately estimated from the BW statistics of enrolment ($Q_{enrol}$)  and test ($Q_{test}$) utterance. Thus, we obtain the overall trial quality as. $Q_{train} \times Q_{test}$. We have used six types of quality measures to improve the performance of ASV system in various duration condition in Table \ref{qmf}.


For a quality fusion function $f_Q(\mathbf{\Lambda})$ with parameters $(\mathbf{\alpha},\beta,\theta)$, the development data $\mathcal{D}$ and an empirical cost function $\hat C((\mathbf{\alpha},\beta,\theta),\mathcal{D})$ are given, the optimal fusion function is obtained by
\begin{equation}
(\mathbf{\alpha}^{\mathrm{dev}},\beta^{\mathrm{dev}},\theta^{\mathrm{dev}})= arg min_{\mathbf{\alpha},\beta,\theta}\hat C(\{\mathbf{\alpha},\beta,\theta\},\mathcal{D}),
\end{equation}

where \emph{decision cost function}, $C$ is defined as,
\begin{equation}
\label{dec_cost}
C_{\mathrm{det}}(\zeta)=C_{\mathrm{miss}}P_{\mathrm{miss}}(\theta)P_{\mathrm{tar}}+C_{\mathrm{fa}}P_{\mathrm{fa}}(\zeta)(1-P_{\mathrm{tar}})
\end{equation}

where $\zeta$ is the threshold, $P_{tar}$ is the prior probability of the target speaker, $C_{\mathrm{miss}}$ is the cost of the miss and $C_{\mathrm{fa}}$ is the cost of the false alarm.

\begin{figure}[t]
\centering
\includegraphics[width=10cm]{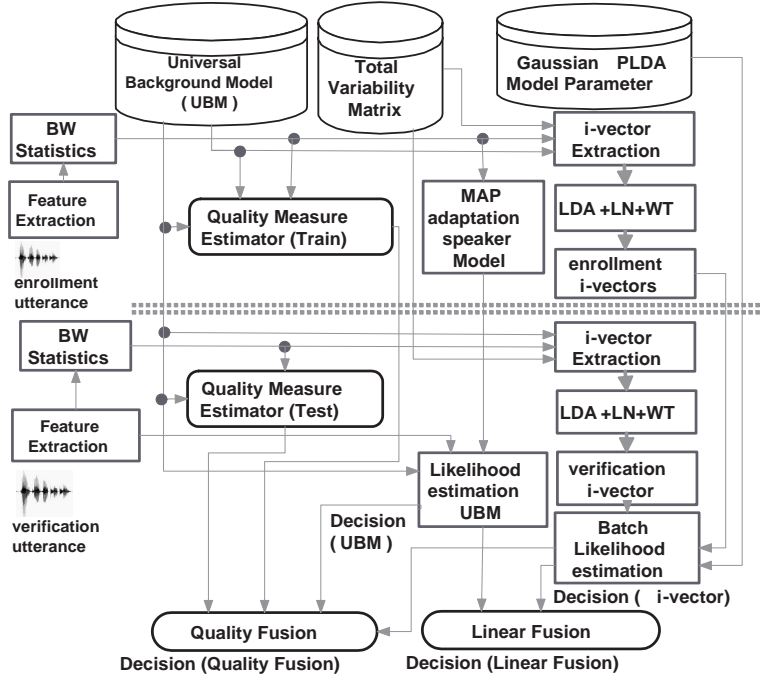}
\caption{Schematic block diagram for proposed speaker verification system showing decisions obtained with both linear fusion and quality-based fusion.}
\label{QivSys}
\end{figure}

A schematic diagram for overall ASV system with linear as well as quality-measure based fusion is shown in Fig.~\ref{QivSys}.

\section{Experimental Setup}\label{Exp_set}

Both GMM-UBM and i-vector based systems use same mel frequency cepstral coefficients (MFCCs)\cite{davis1980comparison} as front-end acoustic features. We extract MFCCs using frame size of 20 ms and frame shift of 10 ms as in~\cite{sahidullah2012design}. The Hamming window is used in MFCC extraction process~\cite{sahidullah2013novel}. The non-speech frames are dropped using energy-based speech activity detector (SAD) \cite{sahidullah2012comparison}. Finally, we performan cepstral mean and variance normalization (CMVN) to remove the convolutive channel effect~\cite{sahidullah2012design}. 19 dimensional MFCC with appended delta and double delta coefficients (57 dimensional) are used throughout the experiments. Gender dependent UBM of 512 mixture components are trained with 10 iterations of EM algorithm. We have used NIST SRE 2004, 2005, 2006 and Switchboard II corpora as development data to estimate UBM, LDA and GPLDA parameters. Total variability subspace of dimension $400$ is chosen for i-vector extractor. We perform LDA on i-vectors to improve the speaker discriminatibility and the dimensions are reduced to $200$. Finally, GPLDA with $150$ eigen-voice space is used for scoring. We estimated the GPLDA parameters with random initialization and 20 iterations of EM algorithm.

\vspace{-3mm}
 \subsection{Experiments and Corpora}
The performance of two major speaker modelling methods and proposed methods were evaluated on NIST SRE 2008 \cite{NIST2008} and NIST SRE 2010 \cite{NIST2010} corpus. NIST 2008 \textit{short2-short-3} and NIST 2010 \textit{core-core} speaker recognition protocol is used as development and evaluation data respectively. Utterance truncated versions of both databases are used for experiments in varying utterance duration condition. Only \textit{telephone-telephone} trials of male speakers from NIST SRE 2008 and NIST SRE 2010 are used in the following experiments.  The summary of the databases used in the experiments are shown in Table~\ref{DatabaseSummary}.

\begin{table}[!t]
  \centering
  \caption{Summary of speech corpora used in the experiments.}

  \begin{tabular}{| l | c | c|}
  \hline
    Specifications & NIST SRE 2008  & NIST SRE 2010\\
    \hline
    \#target model &482  & 489  \\
   \hline
       \#test segments &858  & 351 \\
   \hline
       \#genuine trials & 874 & 353  \\
             \hline
       \#imposter trials &11637  & 13707 \\
   \hline
  \end{tabular}

  \label{DatabaseSummary}
\end{table}

\subsection{Utterance duration and truncation procedure}
In core condition of NIST SRE corpora, the duration of speech segments are long (~2.5 min of speech). In order to conduct experiments in short duration conditions, truncation of speech utterances is done in 2 sec (200 active frames), 5 sec (500 active frames), 10 sec (1000 active frames) and 20 sec (2000 active frames) duration. For truncation of utterances, the prior 500 active speech frames are discarded at the feature level after VAD to avoid phonetic similarity in initial greetings of telephonic conversations which introduces text dependence.
\par
The original utterances from the NIST SRE corpus without any truncation is referred to  as \emph{full} condition in this paper. From the \emph{full} condition features, and features generated from truncated segments, six test sets with different duration conditions in both model and verification segments collection are designed. Fourteen trial conditions are arranged by combining different duration train-test segments and $<$\emph{full}$>$-$<$\emph{duration of test segment}$>$ for both NIST SRE 2008 and 2010. We use the notation `$<$\emph{duration of model segment}$>$-$<$\emph{duration of test segment}$>$ condition,' in which duration is measured in seconds or \emph{full}.

 \subsection{Performance Evaluation Metrics}
 We have evaluated the performance using EER and DCF as performance evaluation metric. The EER is the point on \emph{detection error trade-off} (DET) plot where the probability of false acceptance and probability of false rejection are equal. The DCF is computed by creating a cost function assigning separate weights on false alarm and false rejection followed by computation of threshold where cost function is minimum. The cost function is computed as
     \begin{equation}
     \begin{split}
       C_{\mathrm{Det}}= C_{\mathrm{Miss}} \times P_{\mathrm{Miss|Target}}\times P_{\mathrm{Target}} + C_{\mathrm{FalseAlarm}}
         \times P_{\mathrm{FalseAlarm|NonTarget}} \times ( 1 - P_{\mathrm{Target}})
     \end{split}
     \end{equation}

  The DCF is calculated using the parameter value $C_{\mathrm{Miss}}=10$, $C_{\mathrm{FalseAlarm}}=1$ and $P_{\mathrm{target}}=0.01$ for both databases NIST 2008 and NIST 2010 \cite{NIST2008,NIST2010}. We also report relative improvement for parameter $p$ of system $s_1$ over system $s_2$, calculated as

$ RI_{\mathrm{s_1}}^{\mathrm{p}}=\frac{(p_{\mathrm{s_1}}-p_{\mathrm{s_2}})}{p_{\mathrm{s_2}}}\times 100 \%
$


\begin{table*}
 \centering
\caption{Speaker verification performance on NIST 2008 using GMM-UBM (UBM) and i-vector (TV) based system. The results are shown in terms of EER (in \%) and DCF ($\times$ 100).}
\label{tab:1}       
 \begin{tabular}{|l||c|c|c||c|c|c|}
 \noalign{\smallskip}
 \noalign{\smallskip}
\hline
Train-Test&EER&EER& $RI_{\mathrm{TV}}^{\mathrm{EER}}$&DCF&DCF& $RI_{\mathrm{TV}}^{\mathrm{DCF}}$\\

duration&(UBM)&(TV)&[\%]&(UBM)&(TV)&[\%]\\
\hline\noalign{\smallskip}\hline
\multicolumn{7}{|c|}{\cellcolor{Gray}Truncated training - Truncated testing}\\
\hline\noalign{\smallskip}\hline
2s-2s&\textbf{35.24}& 36.84 & -4.54 & \textbf{9.69} & 9.93 & -2.47\\
5s-5s&25.25 & \textbf{24.37} & 3.49 & 8.89 & \textbf{8.65}& 2.69\\
10s-10s&19.67& \textbf{14.98} & 23.84 & 8.23 & \textbf{6.58} & 20.04\\
20s-20s&16.93 & \textbf{9.72} & 42.58 & 8.19 & \textbf{4.62} & 43.58\\
\hline\noalign{\smallskip}\hline
\multicolumn{7}{|c|}{\cellcolor{Gray}Full training - Truncated testing}\\
\hline\noalign{\smallskip}\hline
Full-2s&21.56 & \textbf{19.67}& 8.76 & \textbf{7.75} & 7.91 & -2.06\\
Full-5s&17.73 & \textbf{13.50} & 23.85 & 7.32 & \textbf{5.99} & 18.16\\
Full-10s&16.66 & \textbf{ 8.78}& 46.93& 6.75 & \textbf{4.50} & 33.33\\
Full-20s&15.52 & \textbf{7.32} & 52.83 & 6.58 & \textbf{3.63} & 41.90\\
\hline\noalign{\smallskip}\hline
Full-Full&14.75 & \textbf{4.86} & 67.05 & 6.23 & \textbf{2.70} & 59.09\\
\hline
\end{tabular}
\end{table*}

\begin{figure}[!t]

\begin{center}
\includegraphics[width=17cm]{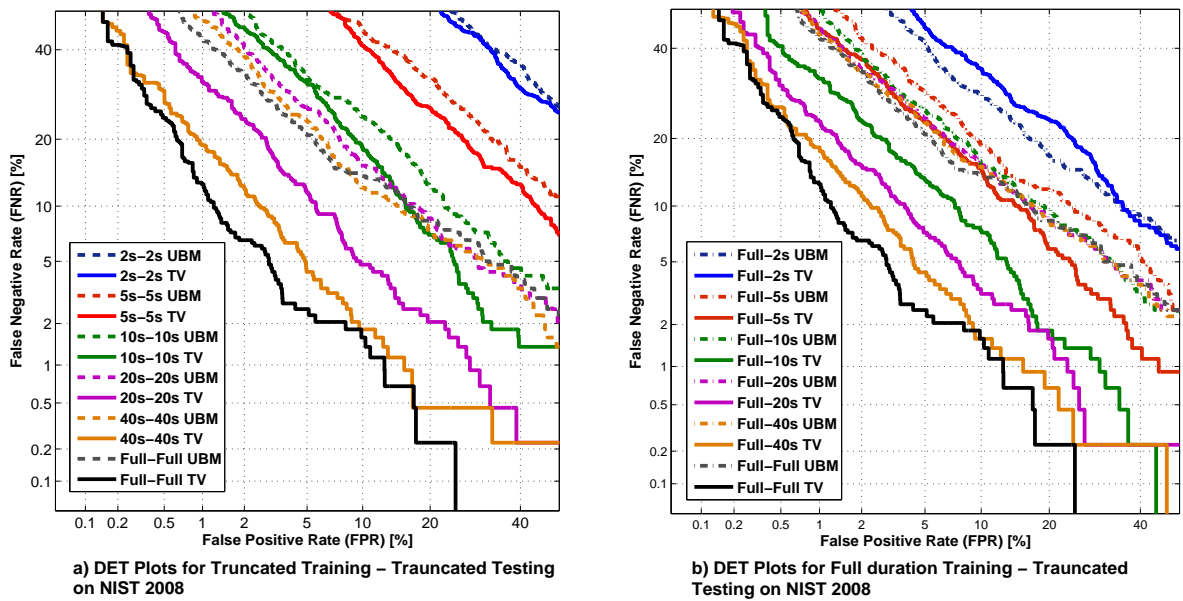}

\end{center}

\caption{Speaker verification DET plot of i-vector (TV) system and UBM system in short2-short3 sub-condition of NIST SRE 2008 corpus.}
\label{DETcmp1}
\end{figure}

\section{Experiments Results and Discussion}
\label{RnD}

\begin{table*}[t]
 \centering
\caption{Results in EER of fusion of GMM-UBM and i-vector based system on NIST 2008 telephone-telephone development corpora }
 \setlength{\tabcolsep}{5pt}
\label{FuTabDevEER}  

 \begin{tabular}{|c|c|c|c|cccccc|}
\noalign{\smallskip}\hline
Train-Test & i-vec & linear & $Q_{tv}$ &$Q_{1}$ & $Q_{2}$ & $Q_{3}$ & $Q_{4}$ & $Q_{5}$ & $Q_{6}$ \\

duration& GPLDA & fusion & & kl-1 &  kl-2 & kl-avg & $\ell_1$ & $\ell_2$ & bh  \\
\hline\hline
\multicolumn{10}{|c|}{\cellcolor{Gray}Truncated training - Truncated testing}\\
\hline\hline
\textbf{2s-2s}  & 36.84 & 32.95 & 33.98 & 31.92 & \textbf{31.80} & 31.75 & 31.92 & 31.88 & \textbf{31.80} \\
\hline
\textbf{5s-2s}  & 30.48 & 30.37 & 28.33 &\textbf{27.23} & 27.45 & 27.57 & 27.45 & 28.48 & 28.72 \\
\textbf{5s-5s}  & 24.37 & 23.11 & 23.17 & 22.99 & 21.62 & \textbf{21.22} & 21.25 & 21.70 & 21.62 \\
\hline
\textbf{10s-2s}  & 26.10 & 25.40 & 23.91 & 23.34 & 23.01 & 23.79 & \textbf{22.88} & 23.21 & 23.11 \\
\textbf{10s-5s}  & 19.56 & 17.50 & 16.84 & 17.16 & 16.93 & 17.16& 16.84 & 16.82 & \textbf{16.73}  \\
\textbf{10s-10s} & 14.98 & 14.53 & 13.95 & 12.92 & 13.72 & 14.30 & 13.84 & \textbf{13.15} & 14.07 \\
\hline
\textbf{20s-2s}   & 24.20 & 21.73 & 22.05 & 20.59 & 20.48 & 20.59 & \textbf{20.18} & 20.48 & 20.36 \\

\textbf{20s-5s}   & 17.29  & 14.94 & 15.95 & 14.94 & 14.94 & 15.07 & 14.94 & 15.23 & \textbf{14.66}  \\

\textbf{20s-10s}  & 12.35 & 11.67 & 11.66 & 11.62 & 11.59 & 11.56 & \textbf{11.55} & 11.67 & 11.63 \\

\textbf{20s-20s}  & 9.72 & 9.83 & 9.83 & \textbf{9.64} & 9.71 & 9.83 & 9.91 & 9.80 & {9.77} \\
\hline\hline
\multicolumn{10}{|c|}{\cellcolor{Gray}Full training - Truncated testing}\\
\hline\hline
\textbf{Full-2s} & 19.67 & 19.10 & 21.51 & \textbf{16.47} & 16.70 & 16.47 & 16.81 & 17.45 & 16.76 \\
\textbf{Full-5s} & 13.50 & 12.70 &13.79 & 11.78 & 11.67 & 11.70 & 11.89 & \textbf{11.83} & 12.12 \\
\textbf{Full-10s} & 9.29 & 9.72 & 14.73 & \textbf{9.15} & 9.24 & 9.27 & 9.09 & 9.26 & 9.49 \\
\textbf{{Full-20s}} & {7.32} & {7.30} & {9.14} & {7.32} & {7.36} & {7.31} & {\textbf{7.26}} & {7.39} & {7.35} \\
\hline\hline
\multicolumn{10}{|c|}{\cellcolor{Gray}Full training - Full testing}\\
\hline\hline
\textbf{{Full-Full}} & {4.86} & {4.91} & {7.29} & {5.20} & {5.20} & {5.14} & {5.06} & {5.16} & {5.14} \\
\hline
\end{tabular}

\end{table*}

\begin{table*}[!ht]
 \centering
\caption{Results in DCF of fusion of GMM-UBM and i-vector based system on NIST 2008 telephone-telephone development corpora }
 \setlength{\tabcolsep}{5pt}
\label{FuTabDevDCF}  

 \begin{tabular}{|c|c|c|c|cccccc|}
\noalign{\smallskip}\hline
Train-Test & i-vec & linear &$Q_{tv}$&$Q_{1}$ & $Q_{2}$ & $Q_{3}$ & $Q_{4}$ & $Q_{5}$ & $Q_{6}$ \\

duration & GPLDA & fusion & & kl-1 &  kl-2 &kl-avg&$\ell_1$& $\ell_2$& bh \\
\hline\hline
\multicolumn{10}{|c|}{\cellcolor{Gray}Truncated training - Truncated testing}\\
\hline\hline
\textbf{2s-2s} & 9.93 & 9.53 & 9.64&\textbf{9.52} &\textbf{ 9.50} & 9.53 & \textbf{9.50} & 9.56 & 9.54 \\
\hline
\textbf{5s-2s} & 9.75 & 9.67 & 9.73 & 8.85 & \textbf{8.87} & 8.87 & 8.92 & 9.13 & 8.89 \\
\textbf{5s-5s} & 8.65 & 8.27 & 8.33 & 8.19 & 8.16 & 8.13 & \textbf{8.19} & 8.24 & 8.24 \\
\hline
\textbf{10s-2s} & 9.23 & 9.05 & 8.54 & 8.27  & 8.28 & \textbf{8.27} & 8.30 & 8.36 & 8.32 \\
\textbf{10s-5s} & 7.80 & 7.34 & 7.30 & 7.15  & \textbf{7.12} & 7.15 & \textbf{7.13} & 7.29 & 7.29 \\
\textbf{10s-10s} & 6.58 & 6.40 & 6.32 & \textbf{5.93} & 6.04 & 6.04 & 5.86 & 6.03 & 5.99 \\
\hline
\textbf{20s-2s} & 8.77 & 8.20 & 8.14 & 7.79 & 7.73 & 7.76 & \textbf{7.79} & 7.79 & 7.80 \\
\textbf{20s-5s} & 6.98 & 6.65 & 6.71 & 6.39 & 6.37 & \textbf{6.34} & 6.34 & 6.47 & 6.46  \\
\textbf{20s-10s} & 5.48 & 5.42 &5.28& 5.36 & 5.24 & 5.26 & \textbf{5.22} & 5.33 & 5.30  \\
{\textbf{20s-20s}} & {4.62} & {4.59} & {4.61} & {4.55} & {4.52} & {4.53} & {\textbf{4.51}} & {4.52} & {\textbf{4.51}}  \\
\hline\hline
\multicolumn{10}{|c|}{\cellcolor{Gray}Full training - Truncated testing}\\
\hline\hline
\textbf{Full-2s} & 7.75 & 7.53 & 7.09 & 7.01 & 7.02 & \textbf{6.99} & 7.01 & 7.09 & 7.09 \\
\textbf{Full-5s} & 5.99 & 5.95 & 6.15 & 5.44 & \textbf{5.36} & 5.42 & 5.19 & 4.93 & 5.57 \\
\textbf{Full-10s} & 4.50 & 4.59 & 6.92 & 4.27 & 4.26 & \textbf{4.25} & 4.23 & 4.36 & 4.32 \\
{\textbf{Full-20s}} & {3.63} & {3.60} &{4.46}& {\textbf{3.61}} & {3.65} & {3.65} & {\textbf{3.61}} & {3.62} & {3.62 } \\
\hline\hline
\multicolumn{10}{|c|}{\cellcolor{Gray}Full training - Full testing}\\
\hline\hline
\textbf{{Full-Full}} & {2.70} & {2.73} & {3.61} & {2.63} & {2.63} & {2.63} & {2.66} & {2.64} & {\textbf{2.61}} \\
\hline

\end{tabular}

\end{table*}

\begin{table*}[!ht]
\renewcommand{\arraystretch}{1.2}
 \centering
\caption{Results in EER of fusion of GMM-UBM and i-vector based system on NIST 2010 telephone-telephone evaluation corpora}
 \setlength{\tabcolsep}{5pt}
\label{FuTabEvalEER}       

 \begin{tabular}{|c|c|c|c|cccccc|}
\noalign{\smallskip}\hline
Train-Test & i-vec & linear &$Q_{tv}$&$Q_{1}$ & $Q_{2}$ & $Q_{3}$ & $Q_{4}$ & $Q_{5}$ & $Q_{6}$ \\

duration& GPLDA & fusion &&  kl-1 &  kl-2 &kl-avg&$\ell_1$& $\ell_2$& bh \\
\hline\hline
\multicolumn{10}{|c|}{\cellcolor{Gray}Truncated training - Truncated testing}\\
\hline\hline
\textbf{2s-2s} & 37.67 & 33.52 & 34.59 & 33.71 & 33.42 & 33.66 & 32.91 & 33.28 & \textbf{32.84}  \\
\hline
\textbf{5s-2s} & 32.01 & 31.44 & 29.13 & 27.44 & \textbf{27.53} & \textbf{27.47} & 27.76 & 28.13 & 27.97 \\
\textbf{5s-5s} & 25.95 & 23.93 & 24.17 & \textbf{21.62} & 22.09 & 21.81 & 21.92 & 22.37 & 22.37 \\
\hline
\textbf{10s-2s} & 28.61 & 27.76 & 24.95 & 24.92 & \textbf{24.65} & 24.65 & 24.54 & 24.92 & 24.92 \\
\textbf{10s-5s} & 20.52 & 18.13 & 17.01 & 17.56 & 16.93 & 16.84 & 16.84 & \textbf{16.82} & \textbf{16.99}  \\
\textbf{10s-10s} & 14.44 & 14.16 & 13.88 & 13.59 & \textbf{13.31} & 13.59 & 13.59 & 13.99 & 13.94  \\
\hline
\textbf{20s-2s} & 24.44 & 21.81 & 21.65 & 21.24 & 21.74 & 21.55 & \textbf{21.12} & 21.51 & 21.50  \\
\textbf{20s-5s} & 16.94 & 14.44 & 15.45 & 14.44 & 14.44 & 14.57 & 14.44 & 14.73 & \textbf{14.16}  \\
\textbf{20s-10s}& 11.28 & 10.19 & 10.08 & 10.07 & \textbf{9.91} & 10.02 & 10.33 & 10.19 & 10.04 \\
\textbf{{20s-20s}}& {8.78} & {8.03} & {9.87} & {8.49} & {8.47} & {8.44} & {8.49} & {\textbf{8.21}} & {8.30} \\
\hline\hline
\multicolumn{10}{|c|}{\cellcolor{Gray}Full training - Truncated testing}\\
\hline\hline
\textbf{Full-2s} & 21.81 & 19.18 & 18.41 & \textbf{17.96} & 18.01 & 17.99 & 18.13 & 18.53 & 18.13 \\
\textbf{Full-5s}  & 12.46 & 11.89 & 13.79 & 11.61 & 11.84 & 11.70 & 11.89 & \textbf{11.44} & 11.99 \\
\textbf{Full-10s} & 7.36 & 7.97 & 13.31 & 7.36 & 7.64 & 7.41 & 7.88 & \textbf{7.08} & 7.59 \\
\textbf{{ Full-20s}} & {5.38} & {5.5} & 7.24 & {\textbf{5.38}} & {5.44} & {\textbf{5.38}} & {5.58} & {5.59} & {5.59}  \\
\hline\hline
\multicolumn{10}{|c|}{\cellcolor{Gray}Full training - Full testing}\\
\hline\hline
\textbf{{Full-Full}} & {2.90} & {2.95} & {5.26} & {3.11} & {3.11} & {3.11} & {3.39} & {3.11} & {3.11} \\
\hline

\end{tabular}

\end{table*}

\begin{table*}[!ht]
\renewcommand{\arraystretch}{1.2}
 \centering
\caption{Results in DCF of fusion of GMM-UBM and i-vector based system on NIST 2010 telephone-telephone evaluation corpora}
 \setlength{\tabcolsep}{5pt}
\label{FuTabEvalDCF}       

 \begin{tabular}{|c|c|c|c|cccccc|}
\noalign{\smallskip}\hline
Train-Test & i-vec & linear & $Q_{tv}$ &$Q_{1}$ & $Q_{2}$ & $Q_{3}$ & $Q_{4}$ & $Q_{5}$ & $Q_{6}$ \\

duration& GPLDA & fusion & &  kl-1 &  kl-2 &kl-avg&$\ell_1$& $\ell_2$& bh \\
\hline\hline
\multicolumn{10}{|c|}{\cellcolor{Gray}Truncated training - Truncated testing}\\
\hline\hline
\textbf{2s-2s} & 9.98 & 9.77 & 9.86 & \textbf{9.74} & 9.77 & \textbf{9.74} & 9.76 & 9.77 & 9.77 \\
\hline
\textbf{5s-2s} & 9.74 & 9.72 & 9.31 & 9.16 & 9.14 & \textbf{9.14} & 9.11 & 9.15 & 9.14  \\
\textbf{5s-5s} & 9.01 & 8.10 & 8.68 & \textbf{8.01} & 8.05 & 8.03 &  8.05 & 8.07 & 8.07  \\
\hline
\textbf{10s-2s} & 9.50 & 9.32 & 8.66 & 8.59 & 8.54 & \textbf{8.50} & 8.56 & 8.48 & 8.48  \\
\textbf{10s-5s} & 7.67 & 7.17 & 7.14 & 7.20 & 7.21 & 7.14 & 7.20 & \textbf{7.04} & \textbf{7.04} \\
\textbf{10s-10s} & 6.52 & 6.29 & 6.18 & 5.97 & 6.17 & 6.05 & 6.05 & \textbf{5.93} & 6.19 \\
\hline
\textbf{20s-2s} & 9.04 & 8.05 & 8.14 & 7.89 & 7.89 & \textbf{7.84} & 7.97 & 7.94 & 7.88  \\
\textbf{20s-5s} & 6.98 & 6.57 & 6.91 & 6.34 & 6.34 & 6.33 & \textbf{6.28} & 6.34 & 6.35  \\
\textbf{20s-10s} & 5.47 & 4.93 & 4.94 & 4.92 & 4.90 & 4.90 & 4.92 & \textbf{4.89} & 4.92 \\
{\textbf{20s-20s}} & {4.11} & {3.85} &{4.89}& {4.03} & {3.96} & {3.94} & {\textbf{3.88}} & {3.94} & {3.97} \\
\hline\hline
\multicolumn{10}{|c|}{\cellcolor{Gray}Full training - Truncated testing}\\
\hline\hline
\textbf{Full-2s} & 8.52 & 7.55 & 7.48 & 7.63 & 7.65 & 7.71 & 7.72 & \textbf{7.61} & \textbf{7.61} \\
\textbf{Full-5s} & 5.47 & 4.92 & 6.15 & 5.14 & 5.13 & 5.14 & 5.19 & \textbf{4.93} & 5.01 \\
\textbf{Full-10s} & 3.69 & 3.67 & 6.00 & 3.65 & 3.62 & \textbf{3.52} & 3.58 & 3.57 & 3.67 \\
{\textbf{Full-20s}} & {2.70} & {2.71} & {3.61}& {2.74} & {2.75} & {2.73} & {\textbf{2.71}} & {2.72} & {\textbf{2.71}} \\
\hline\hline
\multicolumn{10}{|c|}{\cellcolor{Gray}Full training - Full testing}\\
\hline\hline
\textbf{{Full-Full}} & {2.01} & {2.00} & {2.67} & {1.96} & {1.96} & {1.97} & {\textbf{1.95}} & {1.95} & {1.95} \\
\hline

\end{tabular}

\end{table*}

\begin{table*}[!ht]
\renewcommand{\arraystretch}{1.2}
 \centering
\caption{{Results in EER of i-vector with proposed quality metrics on NIST 2008 telephone-telephone evaluation corpora}}
 \setlength{\tabcolsep}{5pt}
\label{FuTab_IV_qual_2008_eer}       

\begin{tabular}{|c|c|cccccc|}
\hline
\begin{tabular}[c]{@{}c@{}}Train - Test \\ Duration\end{tabular} & \begin{tabular}[c]{@{}c@{}}i-vec\\ GPLDA\end{tabular} & Q\_1           & Q\_2          & Q\_2           & Q\_4           & Q\_5           & Q\_6           \\ \hline
\multicolumn{8}{|c|}{\textbf{Truncated Training - Truncated Testing}}                                                                                                                                                         \\ \hline
\textbf{2s - 2s}                                                 & 36.84                                                 & 36.76          & 36.84         & \textbf{36.38} & 36.84          & 36.84          & 36.84          \\ \hline
\textbf{5s - 2s}                                                 & 30.48                                                 & 30.66          & 30.42         & 30.49          & 30.54          & 30.53          & \textbf{30.47} \\
\textbf{5s - 5s}                                                 & 24.37                                                 & \textbf{24.29} & 24.48         & 24.37          & 24.48          & 24.37          & 24.37          \\ \hline
\textbf{10s - 2s}                                                & 26.10                                                 & 26.58          & 26.27         & 26.25          & 26.24          & \textbf{26.10} & 26.19          \\
\textbf{10s - 5s}                                                & 19.56                                                 & 19.56          & 19.01         & 19.31          & \textbf{18.99} & 19.22          & 19.20          \\
\textbf{10s - 10s}                                               & 14.98                                                 & 14.64          & 14.75         & \textbf{14.64} & 14.87          & 14.98          & 14.87          \\ \hline
\textbf{20s - 2s}                                                & 24.20                                                 & 23.94          & 23.90         & 23.91          & \textbf{23.82} & 23.91          & 23.91          \\
\textbf{20s - 5s}                                                &                                                     17.29  & 16.47          & 16.49         & 16.41          & 16.36          & \textbf{16.31} & 16.59          \\
\textbf{20s - 10s}                                               & 12.35                                                 & \textbf{12.14} & 12.31         & 12.12          & 12.35          & 12.24          & 12.24          \\
\textbf{20s - 20s}                                               & 9.72                                                  & \textbf{9.63}  & 9.83          & 9.67           & 9.72           & 9.72           & 9.72           \\ \hline
\multicolumn{8}{|c|}{\textbf{Full Training - Truncated Testing}}                                                                                                                                                              \\ \hline
\textbf{Full - 2s}                                               & 19.67                                                 & \textbf{19.44} & 21.04         & 19.45          & 19.85          & 19.60          & 19.54          \\
\textbf{Full - 5s}                                               & 13.50                                                 & \textbf{13.17} & 13.25         & 13.32          & 13.23          & 13.50          & 13.38          \\
\textbf{Full - 10s}                                              & 9.29                                                  & 9.49           & 9.54          & 9.49           & 9.45           & \textbf{9.28}  & 9.54           \\
\textbf{Full - 20s}                                              & 7.32                                                  & 7.38           & \textbf{7.30} & 7.33           & 7.32           & 7.32           & 7.35           \\ \hline
\end{tabular}
\end{table*}

\begin{table*}[!ht]
\renewcommand{\arraystretch}{1.2}
 \centering
\caption{{Results in DCF of i-vector with proposed quality metrics on NIST 2008 telephone-telephone evaluation corpora}}
 \setlength{\tabcolsep}{5pt}
\label{FuTab_IV_qual_2008_dcf}       

\begin{tabular}{|c|c|cccccc|}
\hline
\begin{tabular}[c]{@{}c@{}}Train - Test \\ Duration\end{tabular} & \begin{tabular}[c]{@{}c@{}}i-vector\\  GPLDA\end{tabular} & Q\_1 & Q\_2 & Q\_2 & Q\_4 & Q\_5 & Q\_6          \\ \hline
\multicolumn{8}{|c|}{\textbf{Truncated Training - Truncated Testing}}                                                                                                           \\ \hline
\textbf{2s - 2s}                                                 & 9.93                                                      & 9.92     &    9.93  &    9.91  &     9.93 & 9.93     &9.93               \\ \hline
\textbf{5s - 2s}                                                 & 9.75                                                      & 9.74 & 9.76 & 9.74 & 9.76 & 9.75 & 9.71          \\
\textbf{5s - 5s}                                                 & 8.65                                                      & 8.72 & 8.70 & 8.71 & 8.67 & 8.65 & 8.64          \\ \hline
\textbf{10s - 2s}                                                & 9.23                                                      & 9.34 & 9.25 & 9.27 & 9.24 & 9.23 & 9.22          \\
\textbf{10s - 5s}                                                & 7.80                                                      & 7.84 & 7.77 & 7.88 & 7.79 & 7.78 & 7.76          \\
\textbf{10s - 10s}                                               & 6.58                                                      & 6.46 & 6.42 & 6.44 & 6.42 & 6.57 & 6.45          \\ \hline
\textbf{20s - 2s}                                                & 8.77                                                      & 8.89 & 8.86 & 8.96 & 8.83 & 8.77 & 8.78          \\
\textbf{20s - 5s}                                                & 6.98                                                      & 7.01 & 6.95 & 6.95 & 6.94 & 6.95 & 6.99          \\
\textbf{20s - 10s}                                               & 5.48                                                      & 5.54 & 5.47 & 5.52 & 5.50 & 5.48 & 5.51          \\
\textbf{20s - 20s}                                               & 4.61                                                      & 4.64 & 4.61 & 4.61 & 4.61 & 4.62 & 4.60          \\ \hline
\multicolumn{8}{|c|}{\textbf{Full Training - Truncated Testing}}                                                                                                                \\ \hline
\textbf{Full - 2s}                                               & 7.75                                                      & 7.88 & 7.89 & 7.87 & 7.91 & 7.91 & 7.87          \\
\textbf{Full - 5s}                                               & 5.99                                                      & 5.93 & 5.92 & 5.91 & 5.91 & 5.99 & 5.94          \\
\textbf{Full - 10s}                                              & 4.50                                                      & 4.38 & 4.35 & 4.34 & 4.34 & 4.50 & 4.42          \\
\textbf{Full - 20s}                                              & 3.63                                                      & 3.60 & 3.54 & 3.62 & 3.51 & 3.63 & \textbf{3.62} \\ \hline
\end{tabular}

\end{table*}

\begin{table*}[!ht]
\renewcommand{\arraystretch}{1.2}
 \centering
\caption{{Results in EER of i-vector with proposed quality metrics on NIST 2010 telephone-telephone evaluation corpora}}
 \setlength{\tabcolsep}{5pt}
\label{FuTab_IV_qual_2010_eer}       

\begin{tabular}{|c|c|cccccc|}
\hline
\begin{tabular}[c]{@{}c@{}}Train - Test \\ Duration\end{tabular} & \begin{tabular}[c]{@{}c@{}}i-vec\\ GPLDA\end{tabular} & Q\_1           & Q\_2          & Q\_2           & Q\_4           & Q\_5           & Q\_6           \\ \hline
\multicolumn{8}{|c|}{\textbf{Truncated Training - Truncated Testing}}                                                                                                                                                         \\ \hline
\textbf{2s - 2s}                                                 & 37.67                                                 & 37.67          & 37.67         & \textbf{38.41} & 37.67          & 37.67          & 37.67          \\ \hline
\textbf{5s - 2s}                                                 & 32.01                                                 & 32.17          & 32.35         & 32.57          & 32.57          & 32.01          & \textbf{32.01} \\
\textbf{5s - 5s}                                                 & 25.95                                                 & 25.77          & 25.77         & 26.01          & \textbf{25.54} & 25.91          & 26.00          \\ \hline
\textbf{10s - 2s}                                                & 28.61                                                 & 28.32          & 28.51         & 28.32          & 28.32          & \textbf{28.54} & 28.39          \\
\textbf{10s - 5s}                                                & 20.52                                                 & 20.39          & 20.46         & 20.24          & \textbf{20.24} & 20.58          & 20.67          \\
\textbf{10s - 10s}                                               & 14.44                                                 & 14.34          & 14.16         & \textbf{14.16} & 14.16          & 14.44          & 14.32          \\ \hline
\textbf{20s - 2s}                                                & 24.44                                                 & 25.49          & 24.57         & 24.36          & \textbf{24.39} & 24.44          & 24.07          \\
\textbf{20s - 5s}                                                & 16.94                                                 & 16.43          & 16.02         & 16.37          & 16.14          & \textbf{16.43} & 15.86          \\
\textbf{20s - 10s}                                               & 11.28                                                 & \textbf{11.22} & 11.32         & 11.30          & 11.20          & 11.29          & 11.22          \\
\textbf{20s - 20s}                                               & 8.21                                                  & \textbf{8.49}  & 8.49          & 8.50           & 8.49           & 8.78           & 8.49           \\ \hline
\multicolumn{8}{|c|}{\textbf{Full Training - Truncated Testing}}                                                                                                                                                              \\ \hline
\textbf{Full - 2s}                                               & 21.81                                                 & \textbf{21.24} & 21.26         & 21.04          & 21.52          & 21.81          & 21.81          \\
\textbf{Full - 5s}                                               & 12.46                                                 & \textbf{12.18} & 12.33         & 12.18          & 12.46          & 12.46          & 12.62          \\
\textbf{Full - 10s}                                              & 7.36                                                  & 7.41           & 7.41          & 7.36           & 7.37           & \textbf{7.36}  & 7.42           \\
\textbf{Full - 20s}                                              & 5.38                                                  & 5.38           & \textbf{5.38} & 5.38           & 5.38           & 5.38           & 5.38           \\ \hline
\end{tabular}

\end{table*}

\begin{table*}[!ht]
\renewcommand{\arraystretch}{1.2}
 \centering
\caption{{Results in DCF of i-vector with proposed quality metrics on NIST 2010 telephone-telephone evaluation corpora}}
 \setlength{\tabcolsep}{5pt}
\label{FuTab_IV_qual_2010_dcf}       
\begin{tabular}{|c|c|cccccc|}
\hline
\begin{tabular}[c]{@{}c@{}}Train - Test \\ Duration\end{tabular} & \begin{tabular}[c]{@{}c@{}}i-vector\\  GPLDA\end{tabular} & Q\_1          & Q\_2          & Q\_2          & Q\_4          & Q\_5          & Q\_6          \\ \hline
\multicolumn{8}{|c|}{\textbf{Truncated Training - Truncated Testing}}                                                                                                                                                        \\ \hline
\textbf{2s - 2s}                                                 & 9.98                                                      & 9.99          & 9.99          & 9.99          & 9.99          & \textbf{9.98} & 9.99          \\ \hline
\textbf{5s - 2s}                                                 & 9.74                                                      & 9.73          & \textbf{9.72} & 9.74          & 9..75         & 9.74          & 9.71          \\
\textbf{5s - 5s}                                                 & 9.01                                                      & \textbf{8.98} & 9.02          & 8.99          & 9.02          & 9.01          & 9.01          \\ \hline
\textbf{10s - 2s}                                                & 9.50                                                      & 9.72          & 9.52          & 9.53          & 9.51          & 9.51          & 9.52          \\
\textbf{10s - 5s}                                                & 7.67                                                      & 7.89          & 7.73          & 7.93          & 7.86          & 7.65          & 7.73          \\
\textbf{10s - 10s}                                               & 6.52                                                      & 6.59          & 6.54          & 6.57          & 6.53          & 6.52          & 6.60          \\ \hline
\textbf{20s - 2s}                                                & 9.04                                                      & \textbf{9.02} & \textbf{9.02} & 9.17          & \textbf{9.02} & 9.04          & 9.03          \\
\textbf{20s - 5s}                                                & 6.98                                                      & 7.07          & 7.01          & 7.04          & 6.99          & \textbf{6.97} & 7.03          \\
\textbf{20s - 10s}                                               & 5.47                                                      & \textbf{5.43} & 5.44          & 5.46          & 5.46          & 5.46          & 5.44          \\
\textbf{20s - 20s}                                               & 4.11                                                      & 4.12          & 4.12          & 4.11          & \textbf{4.10} & 4.11          & 4.12          \\ \hline
\multicolumn{8}{|c|}{\textbf{Full Training - Truncated Testing}}                                                                                                                                                             \\ \hline
\textbf{Full - 2s}                                               & 8.52                                                      & 8.58          & 8.62          & 8.63          & 8.57          & 8.53          & \textbf{8.41} \\
\textbf{Full - 5s}                                               & 5.47                                                      & 5.37          & 5.39          & 5.44          & 5.51          & 5.47          & \textbf{5.32} \\
\textbf{Full - 10s}                                              & 3.69                                                      & 3.76          & 3.82          & 3.80          & 3.86          & 3.70          & 3.75          \\
\textbf{Full - 20s}                                              & 2.71                                                      & 2.72          & 2.73          & \textbf{2.70} & 2.71          & \textbf{2.70} & 2.73          \\ \hline
\end{tabular}
\end{table*}


\subsection{Baseline Performance}
\label{PnC}
Initially, we have investigated the performance of state-of-the-art i-vector and classical GMM-UBM based ASV system under various duration condition. The experiments are executed on male subset of both NIST 2008 \emph{short2-short3} corpus. Comparison of performance is accomplished in eleven different duration conditions separately.

The results of the experiments reported in Table \ref{tab:1}.  Fig.\ref{DETcmp1}
exhibits a systematic comparative study. Table \ref{tab:1}  depicts the relative performance improvement of i-vector based system over GMM-UBM based system i.e. $RI_{\mathrm{TV}}^{\mathrm{EER}}$ and $RI_{\mathrm{TV}}^{\mathrm{DCF}}$ decreases monotonically with the reduction in utterance duration.
   Table \ref{tab:1} also shows that i-vector based system worked better than GMM-UBM for longer utterances but both the system performance falls on duration as small as \emph{2 sec, 5 sec} etc.  We also observed that in case very similar to real-time requirements i.e., very short duration utterances, specially in \textit{full duration training - 2 sec testing} and \textit{2 sec training - 2 sec testing}, GMM-UBM based system showed comparable or even better performance over i-vector based system.
   \par
 The above observations invoke to fuse GMM-UBM and i-vector system which should combine the advantage of both the systems. This is also provides opportunity to include the proposed quality metric as an additional information. Fusion parameters are estimated on NIST 2008 \emph{short2-short3} corpus and validated the same on NIST 2010 \emph{core-core} task for the sake of generality. Quality measure are incorporated in the ASV system to support the fusion device. Table \ref{FuTabDevEER}, \ref{FuTabDevDCF}, \ref{FuTabEvalEER} and \ref{FuTabEvalDCF} represents the results of both linear fusion and quality measure based fusion. Table \ref{FuTabDevEER} and \ref{FuTabEvalEER}  represents the results in terms of EER whereas Table \ref{FuTabDevDCF} and \ref{FuTabEvalDCF} represents in terms of DCF values on development data and evaluation data respectively. In all these tables, we have shown the results for 15  different duration conditions including the \emph{Full-Full} scenario.


\begin{table}[t]
 \caption{Mathematical expressions of duration based quality measures as proposed in~\cite{mandasari2013quality}. Here, $d_m$ and $d_t$ denote the duration of enrolment and test segment where $d_c$ and $k$ are additional parameters.}
 \centerline{
 \setlength{\tabcolsep}{5pt}
\label{Dqmf}       
  \begin{tabular}{lc}
\hline\noalign{\smallskip}\noalign{\smallskip}
 Quality Measure & Additional parameter\\
\noalign{\smallskip}\hline\noalign{\smallskip}\noalign{\smallskip}
 $Q_{dur1}(\tilde{N}_s)=k|\log\frac{d_m}{d_t}|$ & $k$\\
\noalign{\smallskip}\noalign{\smallskip}
 $Q_{dur2}(\tilde{N}_s)=k\log^{2}\frac{d_m}{d_t}$ & $k$\\
\noalign{\smallskip}\noalign{\smallskip}
 $Q_{dur3}(\tilde{N}_s)=k\log\frac{d_m}{d_c}\log\frac{d_c}{d_t}$ & $k, d_c$\\
\noalign{\smallskip}\noalign{\smallskip}
\noalign{\smallskip}\hline
\end{tabular}
}

\end{table}

\begin{table*}[ht]
\renewcommand{\arraystretch}{1.2}
 \centering
\caption{{Comparison of performance of i-vector and GMM-UBM and their linear fusion based ASV system in NIST 2008 telephone-telephone evaluation corpora with Randomized short duration varying between \emph{(2~sec-20~sec)}.}}
 \setlength{\tabcolsep}{5pt}
\label{Rand_base}       

 \begin{tabular}{|c|c|c|c|}
\noalign{\smallskip}\hline

Perf. & i-vec &GMM-UBM&Linear Fusion\\
Metric&GPLDA & & i-vec+GMM-UBM\\
\hline\hline
\multicolumn{4}{|c|}{\cellcolor{Gray}NIST SRE 2008}\\
\hline\hline
EER& 17.41 & 27.68 & 17.24 \\
DCF& 6.83 & 8.67 & 6.71 \\
\hline\hline
\multicolumn{4}{|c|}{\cellcolor{Gray}NIST SRE 2010}\\
\hline\hline
EER& 17.93 & 27.64 & 17.28  \\
DCF & 7.09 & 8.40 & 6.94 \\

\hline

\end{tabular}
\end{table*}

\begin{table*}[ht]
\renewcommand{\arraystretch}{1.2}
 \centering
\caption{{Comparison of performance of proposed Quality metrics in NIST 2008 telephone-telephone evaluation corpora with randomized short duration varying between \emph{(2~sec-20~sec)}.}}
 \setlength{\tabcolsep}{5pt}
\label{Rand}       

 \begin{tabular}{|c|c|ccc|c|cccccc|}
\noalign{\smallskip}\hline

&Baseline& \multicolumn{4}{c|}{{Existing Quality Metrics}} & \multicolumn{6}{c|}{{Proposed Quality Metrics}}\\
\hline
&Linear& \multicolumn{3}{c|}{{Duration}} &{Uncertainty}& \multicolumn{6}{c|}{{BW Statistics}}\\
&Fusion& \multicolumn{3}{c|}{{Based}} & {Based}& \multicolumn{6}{c|}{{Based}}\\
\hline\hline
Perf. & i-vec + &$Q_{dur1}$&$Q_{dur2}$&$Q_{dur3}$&$Q_{tv}$&$Q_{1}$ & $Q_{2}$ & $Q_{3}$ & $Q_{4}$ & $Q_{5}$ & $Q_{6}$\\
Metric&GMM-UBM & &&&&  kl-1 &  kl-2 &kl-avg&$\ell_1$& $\ell_2$& bh \\
\hline\hline
\multicolumn{12}{|c|}{\cellcolor{Gray}NIST SRE 2008}\\
\hline\hline
EER& 17.24 & 17.03 & 17.28 & 17.04 & 24.02 & 16.57 & 16.47 & 16.59 & \textbf{16.36} & 17.04 & 16.80\\
DCF& 6.71 & 6.59 & 6.73& 6.59 & 8.59 & \textbf{6.36} & 6.45 & 6.40 &  6.39 & 6.56 & 6.51  \\
\hline\hline
\multicolumn{12}{|c|}{\cellcolor{Gray}NIST SRE 2010}\\
\hline\hline
EER& 17.28 & 16.14 & 15.86 & \textbf{15.58} & 25.77 & \textbf{15.58} & \textbf{15.58} & \textbf{15.58} & \textbf{15.58} & 15.83 & 15.93 \\
DCF & 6.94 & 6.77 & 6.95 & 6.52 & 8.81 & 6.52 & 6.49 & \textbf{6.46} & 6.50 & 6.53 & 6.59\\

\hline

\end{tabular}
\end{table*}

\subsection{Results of proposed ASV technique}

It may be observed from Tables \ref{FuTabDevEER}, \ref{FuTabDevDCF}, \ref{FuTabEvalEER}, \ref{FuTabEvalDCF}, that the fusion gives higher improvement in shorter utterances.
Irrespective of enrollment and verification, as the utterance length of speech segments becomes longer, the relative difference between performance of GMM-UBM and i-vector system increased considerably.
The fusion based system evolved to be more effective in cases where performance of GMM-UBM and i-vector based system is more comparable i.e., in short utterance cases. The relative boost of the linear fusion based system over i-vector GPLDA based baseline ASV system showed high values up-to $10\%-15\%$ in cases like \emph{full-2~sec, full-5~sec, 20~sec-5~sec, 2~sec-2~sec}, etc. This advocates more potential of the fusion based system
 in real-world scenario.

\begin{figure}[t]
\centering
\includegraphics[width=15cm]{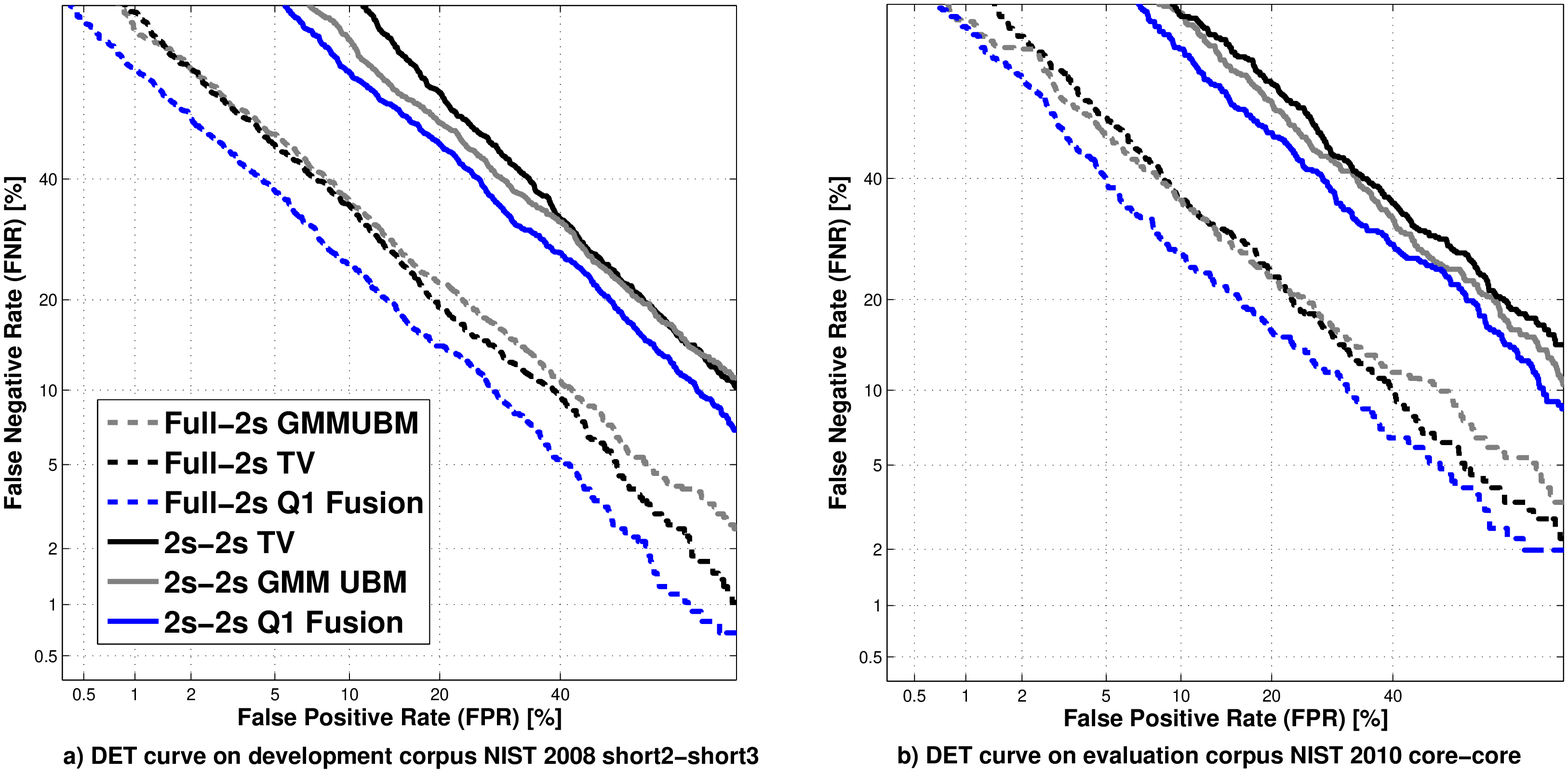}
\caption{DET plot of \emph{Full duration training-2~sec testing} and \emph{2~sec training-2~sec testing 2~sec} on NIST 2008 and NIST 2010 corpora}
\label{DET2}

\end{figure}

Quality measures of speech signals are used to support the fusion ASV system further. Inclusion of proposed quality metrics derived from BW statistics in the proposed system showed significant improvement in performance. Here quality measures are proposed in such a way that it requires almost negligible additional computation and no additional parameters estimation. Performance measures of fusion method using six types of quality measure are shown in Table \ref{FuTabDevEER}, \ref{FuTabDevDCF}, \ref{FuTabEvalEER} and \ref{FuTabEvalDCF}. DET curve showing speaker verification performance of the i-vector baseline system, GMM-UBM based system and proposed quality fusion based system is presented in Fig. \ref{DET2}.



The performance of ASV system with proposed quality metrics improved irrespective of development and evaluation corpora and utterance duration condition as well. { It showed up-to $12\%$ relative improvement over the linear fusion (GMM-UBM+i-vector) based ASV system in conditions like \emph{full-10s, 10s-2s, 5s-2s} etc.} These conditions are more close to desirable real-time requirements of ASV systems which encourages to find implementations of proposed system. Indication of similar improvements both in development and evaluation corpus, shown in Tables \ref{FuTabDevEER}, \ref{FuTabDevDCF}, \ref{FuTabEvalEER} and \ref{FuTabEvalDCF} respectively, authenticates generality of the proposed system. Consistent improvement of accuracy of the ASV system in various duration and databases established relevance of the proposed quality measures based on intermediate statistics. The system is more suitable when the duration speech utterances are limited, especially when it is trained with long enrollment data and tested with very short duration of speech.

\subsection{{Comparison with uncertainty based quality metric}}
A comparison of performance of the proposed quality metrics and an i-vector uncertainty based metric $Q_{tv}$ as in \cite{poorjam2016incorporating},  is accommodated in the aforementioned Tables for different duration condition. The quality measure reflects the duration variability in data as the main source of uncertainty in i-vectors since it has a high correlation with utterance duration.
The posterior distribution of i-vector $\mathbf{y}$ is Gaussian with the following covariance matrix\citep{kenny2008study}
\begin{equation}
\mathbf{y}_{\Sigma}=(\mathbf{I}+ \mathbf{\Phi}^{\top} \mathbf{\Sigma}^{-1}\mathbf{N\Phi)^{-1}}
\end{equation}
  Here the quality measure $Q_{tv}(\mathbf{y}_{\Sigma})$ is calculated as
\begin{equation}
Q_{tv}(\mathbf{y}_{\Sigma}) = \frac{1}{\mathrm{trace}(\mathbf{y}_{\Sigma})}
\end{equation}
Here, in the experiments, truncation of speech utterances are done in fixed durations. Hence, the duration based quality metric, as used for calibration of ASV scores in \cite{mandasari2013quality}, becomes non-functional. The recognition performance with duration based quality metric are compared later in the experiments where the enrollment/verification segment duration are randomized.

\subsection{{Results of quality metrics with only i-vector based system}}

 The proposed quality metrics can also be applied for calibration of the stand-alone classifier based ASV system. The recognition scores of i-vector based system can be calibrated using the proposed quality metrics computed from the speech utterances of the corresponding trials. We have conducted separate experiments to observe the performance of the quality metrics in single classifier based system. We have incorporated the quality metric with the recognition score of i-vector based ASV system using the Eq. \ref{qeqn_ivec}. The performance on NIST 2008 is reported in Table \ref{FuTab_IV_qual_2008_eer} and  \ref{FuTab_IV_qual_2008_dcf} (in \%EER and DCF $\times$ 100 respectively). Whereas, the results on NIST 2010 is presented in Table \ref{FuTab_IV_qual_2010_eer} and  \ref{FuTab_IV_qual_2010_dcf} (in \%EER and DCF $\times$ 100 respectively). The performance metrics as depicted in Tables \ref{FuTab_IV_qual_2008_eer}, \ref{FuTab_IV_qual_2008_dcf}, \ref{FuTab_IV_qual_2010_eer}, \ref{FuTab_IV_qual_2010_dcf}, indicate that the quality metrics showed some improvement over the stand-alone i-vector based ASV system. The results shows marginal improvements with the long and very short duration like Full, 20sec, 2 sec etc. However, the experiments yielded better result in durations like 5 sec, 10 sec etc.

\subsection{Experiments and Results in Mixed duration}
  The evaluation of proposed quality metrics are further extended to more challenging variable duration conditions. Both the enrollment and verification segments are truncated with random duration within a range of \emph{2 sec - 20 sec}. Here the results of the proposed methods are compared with duration-based existing quality metrics as used for calibration in~\cite{mandasari2013quality}. We have used three types of duration based quality metrics ($Q_{dur1}, Q_{dur2}, Q_{dur3}$), as shown in Table~\ref{Dqmf}. The quality measures $Q_{dur1}, Q_{dur2}, Q_{dur3}$ denote function related to the duration of model segment, $d_m$ and the duration of test segment, $d_t$. The value of $d_c$ is kept at 20 sec for the experiments. The fusion parameters ($\alpha, \beta, \theta$) as shown in equation \ref{qeqn}  are estimated from the development set using NIST SRE 2008 and applied for the evaluation with NIST SRE 2010. Table~\ref{Rand} and \ref{Rand_base} present the results on this randomized duration condition for both NIST 2008 and NIST 2010 corpora. In Table~\ref{Rand_base}, we have compared the three baseline systems namely i-vector (GPLDA), GMM-UBM and linear fusion of the two in randomized duration condition for both train and test utterances. However, in Table~\ref{Rand}, the performance of the existing duration based and uncertainty based quality metrics are compared with proposed quality metrics, taking linear fusion of i-vector and GMM-UBM system as baseline.  The results show a consistent improvement for the proposed methods. Further, the proposed methods outperform the duration-based and uncertainty based quality measures in both databases.

\section{Conclusion and Future Scopes}
\label{conclude}
{In this work, we have introduced new quality measures for improving the speaker recognition performance under short duration conditions. We derive the quality measures using Baum-Welch sufficient statistics which are used for computing i-vector representation. We demonstrate that the dissimilarity between the normalized zero-order Baum-Welch statistics and the weights of universal background model (UBM) is associated with the speech duration. We formulate the quality measures based on the normalized zero-order Baum-Welch statistics and UBM weights. This quality measure estimation method does not require additional parameter estimation as they directly derived from already estimated parameters. The proposed quality measures of speech are incorporated as side information in fusion-based ASV system with i-vector and \emph{Gaussian mixture model-Universal background model} (GMM-UBM) system as two subsystems. The score fusion with proposed quality measures substantially enhanced the ASV performance, especially for short utterance. We observed up to $12.63\%$ relative improvement over linear fusion based ASV system. The performance is also considerably better than the performance obtained with existing speech duration and uncertainty based quality measures. Even though we have observed considerable improvement with the distance-based proposed quality measures, we do not observe any clear indication on which quality measure distance function is most appropriate. This also opens up the possibility of further optimizing the distance measures for quality estimation.}

{In this work, we have considered GMM-based i-vectors. As an extension of this study, similar investigations on quality measure fusion can be made with the latest DNN-ASR-based i-vector system. The proposed method can also be explored for ASV system fusions including x-vector system as a subsystem. Moreover, the distance-based quality measures implicitly represent the acoustic variations in the speech utterance with respect to the mean of UBM or the distribution of acoustic space. Therefore, it would be interesting to explore the general use case of the proposed quality measures where acoustic variability needs to be computed.}

\section{Acknowledgment}
This work is partially supported by Indian Space Research Organization (ISRO), Government Of India. The work of Md Sahidullah is supported by Region Grand Est, France. The authors would like to express their sincere thanks to the anonymous reviewers and the editors for their comments and suggestions, which greatly improved the work in quality and content. We further thank Dr. Tomi Kinnunen (University of Eastern Finland) for his valuable comments on an the earlier version of this work. Finally, the authors would also like to acknowledge Dr. Monisankha Pal (Signal Analysis and Interpretation Laboratory, University of Southern California), Shefali Waldekar (ABSP lab, Dept. of E \& ECE, IIT Kharagpur), and Dipannita Podder (VIP Lab, Dept. of CSE, IIT Kharagpur) for their helpful suggestions in different stages of this work.

\section*{References}
\bibliography{mybibfile}

\begin{thebibliography}{10}
\expandafter\ifx\csname url\endcsname\relax
  \def\url#1{\texttt{#1}}\fi
\expandafter\ifx\csname urlprefix\endcsname\relax\def\urlprefix{URL }\fi
\expandafter\ifx\csname href\endcsname\relax
  \def\href#1#2{#2} \def\path#1{#1}\fi

\bibitem{kinnunen2010overview}
T.~Kinnunen, H.~Li, An overview of text-independent speaker recognition: {F}rom
  features to supervectors, Speech Communication 52~(1) (2010) 12--40.

\bibitem{campbell2009forensic}
J.~P. Campbell, W.~Shen, W.~M. Campbell, R.~Schwartz, J.~Bonastre, D.~Matrouf,
  Forensic speaker recognition, IEEE Signal Processing Magazine 26~(2) (2009)
  95--103.

\bibitem{ICICI}
{ICICI} bank introduces voice recognition for biometric authentication,
  \url{https://www.icicibank.com/aboutus/article.page?identifier=news-icici-bank-introduces-voice-recognition-for-biometric-authentication-20152505124050634},
  accessed: 2019-01-25.

\bibitem{Barclays}
Death of the password?,
  \url{https://www.barclayscorporate.com/insight-and-research/fraud-smart-centre/biometrics.html},
  accessed: 2019-01-25.

\bibitem{INTERPOL}
Interpol’s new software will recognize criminals by their voices,
  \url{https://spectrum.ieee.org/tech-talk/consumer-electronics/audiovideo/interpols-new-automated-platform-will-recognize-criminals-by-their-voice},
  accessed: 2019-01-25.

\bibitem{campbell1997speaker}
J.~P. Campbell~Jr, Speaker recognition: {A} tutorial, Proceedings of the IEEE
  85~(9) (1997) 1437--1462.

\bibitem{dehak2011front}
N.~Dehak, P.~Kenny, R.~Dehak, P.~Dumouchel, P.~Ouellet, Front-end factor
  analysis for speaker verification, IEEE Transactions on Audio, Speech, and
  Language Processing 19~(4) (2011) 788--798.

\bibitem{matvejka2016analysis}
P.~Mat{\v{e}}jka, O.~Glembek, O.~Novotn{\`y}, O.~Plchot, F.~Gr{\'e}zl,
  L.~Burget, J.~H. Cernock{\`y}, Analysis of dnn approaches to speaker
  identification, in: Proc. ICASSP, IEEE, 2016, pp. 5100--5104.

\bibitem{snyder2018xvector}
D.~Snyder, D.~Garcia-Romero, G.~Sell, D.~Povey, S.~Khudanpur, X-vectors: Robust
  {DNN} embeddings for speaker recognition, in: Proc. ICASSP, 2018.

\bibitem{poddar2017speaker}
A.~Poddar, M.~Sahidullah, G.~Saha, Speaker verification with short utterances:
  a review of challenges, trends and opportunities, IET Biometrics 7~(3) (2018)
  91--101.

\bibitem{solewicz2013estimated}
Y.~A. Solewicz, G.~Jardine, T.~Becker, S.~Gfroerer, Estimated intra-speaker
  variability boundaries in forensic speaker recognition casework, Proceedings
  of Biometric Technologies in Forensic Science (BTFS)(Nijmegen) (2013) 31--33.

\bibitem{ming2007robust}
J.~Ming, T.~J. Hazen, J.~R. Glass, D.~A. Reynolds, Robust speaker recognition
  in noisy conditions, IEEE Transactions on Audio, Speech, and Language
  Processing 15~(5) (2007) 1711--1723.

\bibitem{saeidi2016feature}
R.~Saeidi, P.~Alku, T.~B{\"a}ckstr{\"o}m, Feature extraction using power-law
  adjusted linear prediction with application to speaker recognition under
  severe vocal effort mismatch, IEEE/ACM Transactions on Audio, Speech and
  Language Processing 24~(1) (2016) 42--53.

\bibitem{mclaren2016exploring}
M.~McLaren, L.~Ferrer, A.~Lawson, Exploring the role of phonetic bottleneck
  features for speaker and language recognition, in: Proc. ICASSP, IEEE, 2016,
  pp. 5575--5579.

\bibitem{parthasarathy2017study}
S.~Parthasarathy, C.~Zhang, J.~H. Hansen, C.~Busso, A study of speaker
  verification performance with expressive speech, in: Proc. ICASSP, IEEE,
  2017, pp. 5540--5544.

\bibitem{wang2016robust}
D.~Wang, Y.~Zou, J.~Liu, Y.~Huang, A robust {DBN}-vector based speaker
  verification system under channel mismatch conditions, in: 2016 IEEE
  International Conference on Digital Signal Processing (DSP), IEEE, 2016, pp.
  94--98.

\bibitem{vestman2017time}
V.~Vestman, D.~Gowda, M.~Sahidullah, P.~Alku, T.~Kinnunen, Time-varying
  autoregressions for speaker verification in reverberant conditions, in: Proc.
  INTERSPEECH, 2017, pp. 1512--1516.

\bibitem{kanagasundaram2011vector}
A.~Kanagasundaram, R.~Vogt, D.~B. Dean, S.~Sridharan, M.~W. Mason, I-vector
  based speaker recognition on short utterances, in: Proc. INTERSPEECH,
  International Speech Communication Association (ISCA), 2011, pp. 2341--2344.

\bibitem{mandasari2011evaluation}
M.~I. Mandasari, M.~McLaren, D.~A. van Leeuwen, Evaluation of i-vector speaker
  recognition systems for forensic application, in: Proc. INTERSPEECH, 2011,
  pp. 21--24.

\bibitem{kanagasundaram2012plda}
A.~Kanagasundaram, R.~J. Vogt, D.~B. Dean, S.~Sridharan, {PLDA} based speaker
  recognition on short utterances, in: Proc. Odyssey: The Speaker and Language
  Recognition Workshop, ISCA, 2012.

\bibitem{sarkar2012study}
A.~K. Sarkar, D.~Matrouf, P.-M. Bousquet, J.-F. Bonastre, Study of the effect
  of i-vector modeling on short and mismatch utterance duration for speaker
  verification, in: Proc. INTERSPEECH, 2012.

\bibitem{fauve2007influence}
B.~G. Fauve, N.~W. Evans, N.~Pearson, J.-F. Bonastre, J.~S. Mason, Influence of
  task duration in text-independent speaker verification, in: Proc.
  INTERSPEECH, 2007, pp. 794--797.

\bibitem{kajarekar2003modeling}
L.~Ferrer, H.~Bratt, V.~R. Gadde, S.~S. Kajarekar, E.~Shriberg, K.~Sonmez,
  A.~Stolcke, A.~Venkataraman, Modeling duration patterns for speaker
  recognition, in: Proc. EUROSPEECH, 2003.

\bibitem{fauve2008improving}
B.~Fauve, N.~Evans, J.~Mason, Improving the performance of text-independent
  short duration {SVM}-and {GMM}-based speaker verification, in: Proc. Odyssey:
  The Speaker and Language Recognition Workshop, 2008, p.~18.

\bibitem{hasan2013duration}
T.~Hasan, R.~Saeidi, J.~H. Hansen, D.~van Leeuwen, Duration mismatch
  compensation for i-vector based speaker recognition systems, in: Proc.
  ICASSP, IEEE, 2013, pp. 7663--7667.

\bibitem{kanagasundaram2014improving}
A.~Kanagasundaram, D.~Dean, S.~Sridharan, J.~Gonzalez-Dominguez,
  J.~Gonzalez-Rodriguez, D.~Ramos, Improving short utterance i-vector speaker
  verification using utterance variance modelling and compensation techniques,
  Speech Communication 59 (2014) 69--82.

\bibitem{mandasari2015quality}
M.~I. Mandasari, R.~Saeidi, D.~A. van Leeuwen, Quality measures based
  calibration with duration and noise dependency for speaker recognition,
  Speech Communication 72 (2015) 126--137.

\bibitem{mandasari2013quality}
M.~I. Mandasari, R.~Saeidi, M.~McLaren, D.~A. van Leeuwen, Quality measure
  functions for calibration of speaker recognition systems in various duration
  conditions, IEEE Transactions on Audio, Speech, and Language Processing
  21~(11) (2013) 2425--2438.

\bibitem{li2016improving}
L.~Li, D.~Wang, C.~Zhang, T.~F. Zheng, Improving short utterance speaker
  recognition by modeling speech unit classes, IEEE/ACM Transactions on Audio,
  Speech, and Language Processing 24~(6) (2016) 1129--1139.

\bibitem{zhang2018text}
C.~Zhang, K.~Koishida, J.~Hansen, Text-independent speaker verification based
  on triplet convolutional neural network embeddings, IEEE/ACM Transactions on
  Audio, Speech and Language Processing 26~(9) (2018) 1633--1644.

\bibitem{guo2018deep}
J.~Guo, N.~Xu, K.~Qian, Y.~Shi, K.~Xu, Y.~Wu, A.~Alwan, Deep neural network
  based i-vector mapping for speaker verification using short utterances,
  Speech Communication 105 (2018) 92--102.

\bibitem{poh2010quality}
N.~Poh, J.~Kittler, T.~Bourlai, Quality-based score normalization with device
  qualitative information for multimodal biometric fusion, IEEE Transactions on
  Systems, Man, and Cybernetics-Part A: Systems and Humans 40~(3) (2010)
  539--554.

\bibitem{poh2005improving}
N.~Poh, S.~Bengio, Improving fusion with margin-derived confidence in biometric
  authentication tasks, in: International Conference on Audio-and Video-Based
  Biometric Person Authentication, Springer, 2005, pp. 474--483.

\bibitem{poh2010multimodal}
N.~Poh, T.~Bourlai, J.~Kittler, A multimodal biometric test bed for
  quality-dependent, cost-sensitive and client-specific score-level fusion
  algorithms, Pattern Recognition 43~(3) (2010) 1094--1105.

\bibitem{poh2012unified}
N.~Poh, J.~Kittler, A unified framework for biometric expert fusion
  incorporating quality measures, IEEE Transactions on Pattern Analysis and
  Machine Intelligence 34~(1) (2012) 3--18.

\bibitem{fierrez2005discriminative}
J.~Fierrez-Aguilar, J.~Ortega-Garcia, J.~Gonzalez-Rodriguez, J.~Bigun,
  Discriminative multimodal biometric authentication based on quality measures,
  Pattern recognition 38~(5) (2005) 777--779.

\bibitem{grother2007performance}
P.~Grother, E.~Tabassi, Performance of biometric quality measures, IEEE
  Transactions on Pattern Analysis and Machine Intelligence 29~(4) (2007)
  531--543.

\bibitem{garcia2006using}
D.~Garcia-Romero, J.~Fierrez-Aguilar, J.~Gonzalez-Rodriguez, J.~Ortega-Garcia,
  Using quality measures for multilevel speaker recognition, Computer Speech \&
  Language 20~(2-3) (2006) 192--209.

\bibitem{garcia2004use}
D.~Garcia-Romero, J.~Fi{\'e}rrez-Aguilar, J.~Gonzalez-Rodriguez,
  J.~Ortega-Garcia, On the use of quality measures for text-independent speaker
  recognition, in: Proc. Odyssey: The Speaker and Language Recognition
  Workshop, 2004.

\bibitem{hasan2013crss}
T.~Hasan, S.~O. Sadjadi, G.~Liu, N.~Shokouhi, H.~Bo{\v{r}}il, J.~H. Hansen,
  {CRSS} systems for 2012 {NIST} speaker recognition evaluation, in: Proc.
  ICASSP, 2013, pp. 6783--6787.

\bibitem{harriero2009analysis}
A.~Harriero, D.~Ramos, J.~Gonzalez-Rodriguez, J.~Fierrez, Analysis of the
  utility of classical and novel speech quality measures for speaker
  verification, in: International Conference on Biometrics, Springer, 2009, pp.
  434--442.

\bibitem{alonso2012quality}
F.~Alonso-Fernandez, J.~Fierrez, J.~Ortega-Garcia, Quality measures in
  biometric systems, IEEE Security \& Privacy 10~(6) (2012) 52--62.

\bibitem{chibelushi2002review}
C.~C. Chibelushi, F.~Deravi, J.~S. Mason, A review of speech-based bimodal
  recognition, IEEE Transactions on Multimedia 4~(1) (2002) 23--37.

\bibitem{kittler2007quality}
J.~Kittler, N.~Poh, O.~Fatukasi, K.~Messer, K.~Kryszczuk, J.~Richiardi,
  A.~Drygajlo, Quality dependent fusion of intramodal and multimodal biometric
  experts, in: Defense and Security Symposium, International Society for Optics
  and Photonics, 2007, pp. 653903--653903.

\bibitem{arnab2017icapr}
A.~Poddar, M.~Sahidullah, G.~Saha, Novel quality metric for duration
  variability compensation in speaker verification, in: Proc. Ninth
  International Conference on Advances in Pattern Recognition (ICAPR-2017),
  2017.

\bibitem{poddar2018improved}
A.~Poddar, M.~Sahidullah, G.~Saha, Improved i-vector extraction technique for
  speaker verification with short utterances, International Journal of Speech
  Technology 21~(3) (2018) 473--488.

\bibitem{reynolds2000speaker}
D.~A. Reynolds, T.~F. Quatieri, R.~B. Dunn, Speaker verification using adapted
  {G}aussian mixture models, Digital Signal Processing 10~(1) (2000) 19--41.

\bibitem{campbell2006svm}
W.~M. Campbell, D.~E. Sturim, D.~A. Reynolds, A.~Solomonoff, {SVM} based
  speaker verification using a {GMM} supervector kernel and {NAP} variability
  compensation, in: Proc. ICASSP, Vol.~1, IEEE, 2006, pp. I--I.

\bibitem{kenny2007joint}
P.~Kenny, G.~Boulianne, P.~Ouellet, P.~Dumouchel, Joint factor analysis versus
  eigenchannels in speaker recognition, IEEE Transactions on Audio, Speech, and
  Language Processing 15~(4) (2007) 1435--1447.

\bibitem{kenny2010bayesian}
P.~Kenny, Bayesian speaker verification with heavy-tailed priors, in: Proc.
  Odyssey: The Speaker and Language Recognition Workshop, 2010, p.~14.

\bibitem{li2016feature}
W.~Li, T.~Fu, H.~You, J.~Zhu, N.~Chen, Feature sparsity analysis for i-vector
  based speaker verification, Speech Communication 80 (2016) 60--70.

\bibitem{poorjam2016incorporating}
A.~H. Poorjam, R.~Saeidi, T.~Kinnunen, V.~Hautam{\"a}ki, Incorporating
  uncertainty as a quality measure in i-vector based language recognition, in:
  Proc. Odyssey: The Speaker and Language Recognition Workshop, ISCA, 2016, pp.
  74--80.

\bibitem{ferrer2005class}
L.~Ferrer, M.~K. S{\"o}nmez, S.~S. Kajarekar, Class-dependent score combination
  for speaker recognition, in: Proc. INTERSPEECH, 2005, pp. 2173--2176.

\bibitem{doddington2000nist}
G.~R. Doddington, M.~A. Przybocki, A.~F. Martin, D.~A. Reynolds, The {NIST}
  speaker recognition evaluation--overview, methodology, systems, results,
  perspective, Speech Communication 31~(2) (2000) 225--254.

\bibitem{bigun2003multimodal}
J.~Bigun, J.~Fierrez-Aguilar, J.~Ortega-Garcia, J.~Gonzalez-Rodriguez,
  Multimodal biometric authentication using quality signals in mobile
  communications, in: Proc. 12th International Conference on Image Analysis and
  Processing, 2003.

\bibitem{hautamaki2013sparse}
V.~Hautamaki, T.~Kinnunen, F.~Sedl{\'a}k, K.~A. Lee, B.~Ma, H.~Li, Sparse
  classifier fusion for speaker verification, IEEE Transactions on Audio,
  Speech, and Language Processing 21~(8) (2013) 1622--1631.

\bibitem{bosaris2011}
{BOSARIS} {T}oolkit [software package], available at:
  https://sites.google.com/site/bosaristoolkit.

\bibitem{NIST2008}
The {NIST} year 2008 speaker recognition evaluation plan, tech.rep., NIST.

\bibitem{NIST2010}
The {NIST} year 2010 speaker recognition evaluation plan, tech.rep., NIST.

\bibitem{arnab2015comparison}
A.~Poddar, M.~Sahidullah, G.~Saha, Performance comparison of speaker
  recognition systems in presence of duration variability, in: Proc. 2015
  Annual IEEE India Conference (INDICON), IEEE, 2015, pp. 1--6.

\bibitem{davis1980comparison}
S.~B. Davis, P.~Mermelstein, Comparison of parametric representations for
  monosyllabic word recognition in continuously spoken sentences, IEEE
  Transactions on Acoustics, Speech and Signal Processing 28~(4) (1980)
  357--366.

\bibitem{sahidullah2012design}
M.~Sahidullah, G.~Saha, Design, analysis and experimental evaluation of block
  based transformation in {MFCC} computation for speaker recognition, Speech
  Communication 54~(4) (2012) 543--565.

\bibitem{sahidullah2013novel}
M.~Sahidullah, G.~Saha, A novel windowing technique for efficient computation
  of {MFCC} for speaker recognition, Signal Processing Letters 20~(2) (2013)
  149--152.

\bibitem{sahidullah2012comparison}
M.~Sahidullah, G.~Saha, Comparison of speech activity detection techniques for
  speaker recognition, arXiv preprint arXiv:1210.0297.

\bibitem{kenny2008study}
P.~Kenny, P.~Ouellet, N.~Dehak, V.~Gupta, P.~Dumouchel, A study of interspeaker
  variability in speaker verification, IEEE Transactions on Audio, Speech, and
  Language Processing 16~(5) (2008) 980--988.

\end{thebibliography}

\section*{}

\textbf{Arnab Poddar} received his MS (by research) degree in the area of speech processing and machine learning from the Department of Electronics \& Electrical Communication Engineering, Indian Institute Technology Kharagpur in 2018. He has worked in the research project entitled as \textit{Reduction of False acceptance an rejection in non-cooperative automatic speaker recognition system}, funded by Indian Space Research Organization (ISRO). Prior to that he has worked as a research project person in the project \textit{Development of Optical Character Recognition system on printed Indian Languages} in Computer Vision and Pattern Recognition (CVPR) Unit, Indian Statistical Institute (ISI). He is currently pursuing Ph.D. at Indian Institute of Technology Kharagpur in area of machine learning and computer vision. His research interests include speech \& audio signal processing, image processing, and machine learning.

\vspace{1cm}

\textbf{Md Sahidullah} received his Ph.D. degree in the area of speech processing from the Department of Electronics \& Electrical Communication Engineering, Indian Institute Technology Kharagpur in 2015. Prior to that he obtained the Bachelors of Engineering degree in Electronics and Communication Engineering from Vidyasagar University in 2004 and the Masters of Engineering degree in Computer Science and Engineering (with specialization in Embedded System) from West Bengal University of Technology in 2006. In 2007-2008, he was with Cognizant Technology Solutions India PVT Limited. In 2014-2017, he was a postdoctoral researcher with the School of Computing, University of Eastern Finland. In January 2018, he joined MULTISPEECH team, Inria, France as a post-doctoral researcher where he currently holds a starting research position. His research interest includes robust speaker recognition, voice activity detection and spoofing countermeasures. He is also a co-organizer of two \textit{Automatic Speaker Verification
Spoofing and Countermeasures Challenges}: ASVspoof 2017 and ASVspoof 2019.

\vspace{1cm}

\textbf{Goutam Saha} received his B.Tech. and Ph.D. degrees from the Department of Electronics \& Electrical Communication Engineering, Indian Institute of Technology (IIT) Kharagpur, India in 1990 and 2000, respectively. In between, he served industry for about four years and obtained a five year fellowship from Council of Scientific \& Industrial Research, India. In 2002, he joined IIT Kharagpur as a faculty member where he is currently serving as a Professor. His research interests include analysis of audio and bio signals.

\end{document}